\documentclass[runningheads]{llncs}

\usepackage{eccv}

\usepackage{xcolor,colortbl}
\definecolor{gaincolor}{RGB}{230, 245, 255}
\usepackage{tcolorbox}

\definecolor{gold}{HTML}{BD820B}
\definecolor{silver}{HTML}{909090}
\definecolor{bronze}{HTML}{9A5F26}
\definecolor{lgray}{gray}{0.95}

\definecolor{Gray}{gray}{0.91}
\definecolor{LightCyan}{rgb}{0.82,0.82,1}
\newcolumntype{a}{>{\columncolor{Gray}}c}
\newcolumntype{B}{>{\columncolor{LightCyan}}c}

\usepackage{eccvabbrv}
\usepackage{amssymb}
\usepackage{pifont}
\newcommand{\cmark}{\ding{51}}%
\newcommand{\xmark}{\ding{55}}%

\usepackage{graphicx}
\usepackage{booktabs}
\usepackage{soul}

\usepackage[accsupp]{axessibility}  

\usepackage{hyperref}

\usepackage{orcidlink}
\usepackage{tabularx}
\usepackage{multicol}
\usepackage{multirow}

\begin{document}


\title{Falcon \includegraphics[height=1em]{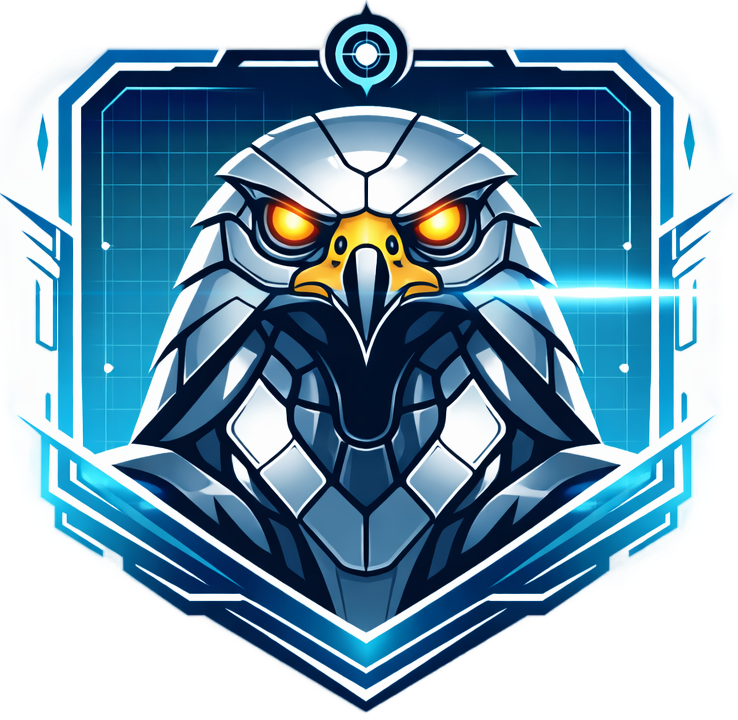}\hspace{0.1em}: Functional Assembly and Language for Compositional Reasoning in X-ray} 

\titlerunning{Falcon}

\author{
Yonathan Michael\orcidlink{0009-0005-8457-294X}\textsuperscript{\dag}\thanks{Equal contribution.}
\and
Mohamad Alansari\orcidlink{0000-0003-2960-2972}\textsuperscript{\dag}$^{*}$
\and
Natnael Takele\orcidlink{0009-0009-3788-5256}\textsuperscript{\dag}
\and
Andreas Henschel\orcidlink{0000-0003-1386-5372}\textsuperscript{\dag\ddag}
\and
Naoufel Werghi\orcidlink{0000-0002-5542-448X}\textsuperscript{\dag\ddag}
}

\authorrunning{Y. ~Michael, M. Alansari et al.}

\institute{
\textsuperscript{\dag}Computer Science Department \quad
\textsuperscript{\ddag}Center of Cyber-physical  Systems (C2PS)\\ Khalifa University\\
\email{\{100053679,100061914,100058082,andreas.henschel,naoufel.werghi\}@ku.ac.ae} \\
\href{https://yonathan-kiflom.github.io/FALCON/page/}{https://yonathan-kiflom.github.io/FALCON/}
}

\maketitle
\vspace{-2em}
\begin{figure*}[h!]
\centering
\includegraphics[width=0.7\linewidth,height=3cm]{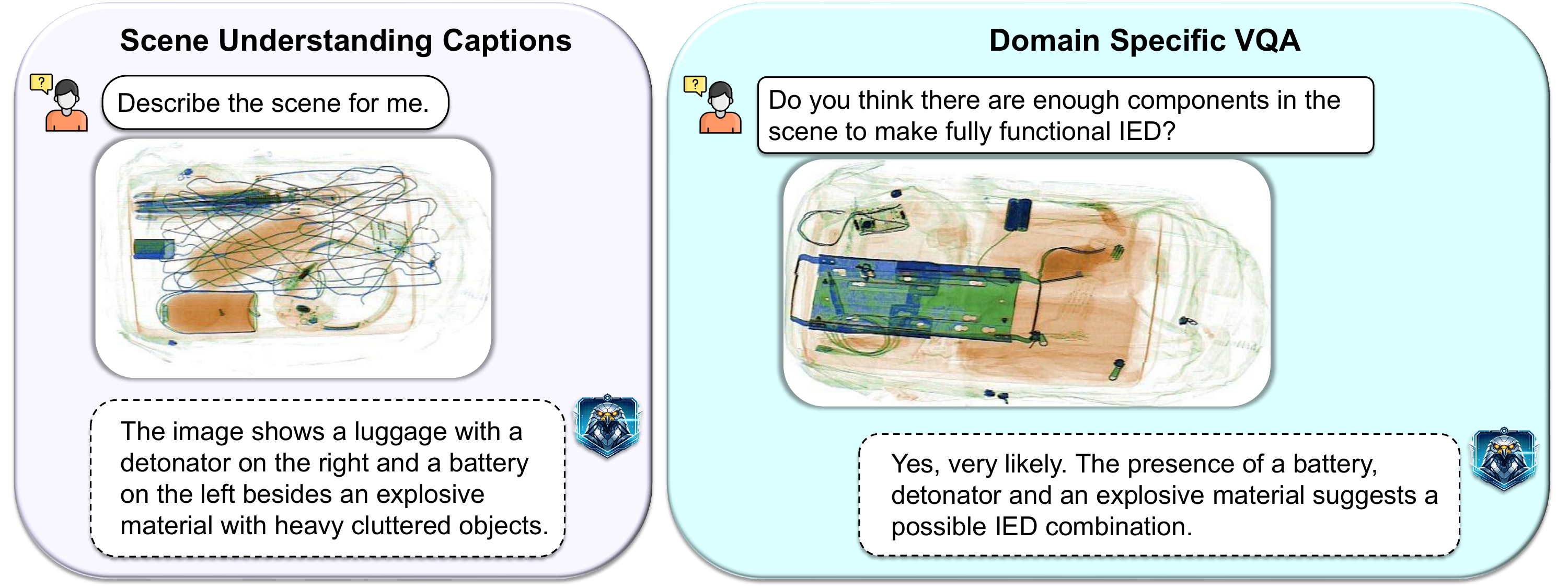}
\includegraphics[width=0.8\linewidth,height=3cm]{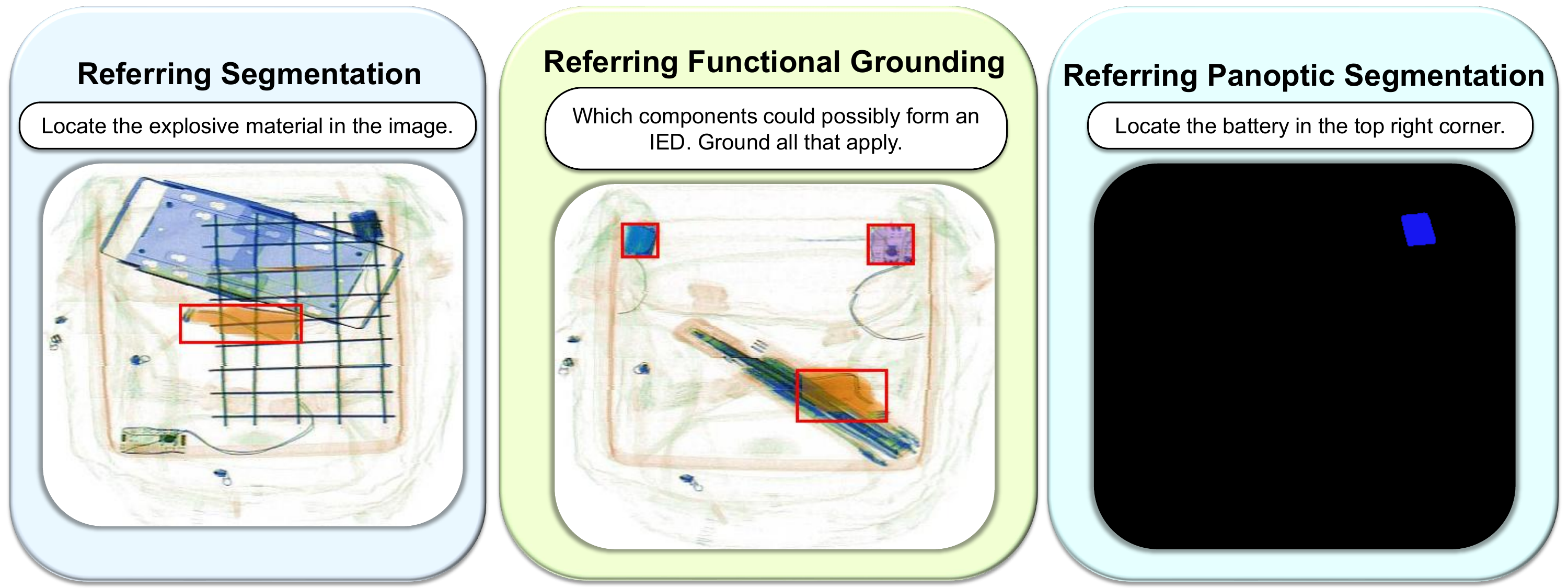}

\includegraphics[width=0.7\linewidth,height=3cm]{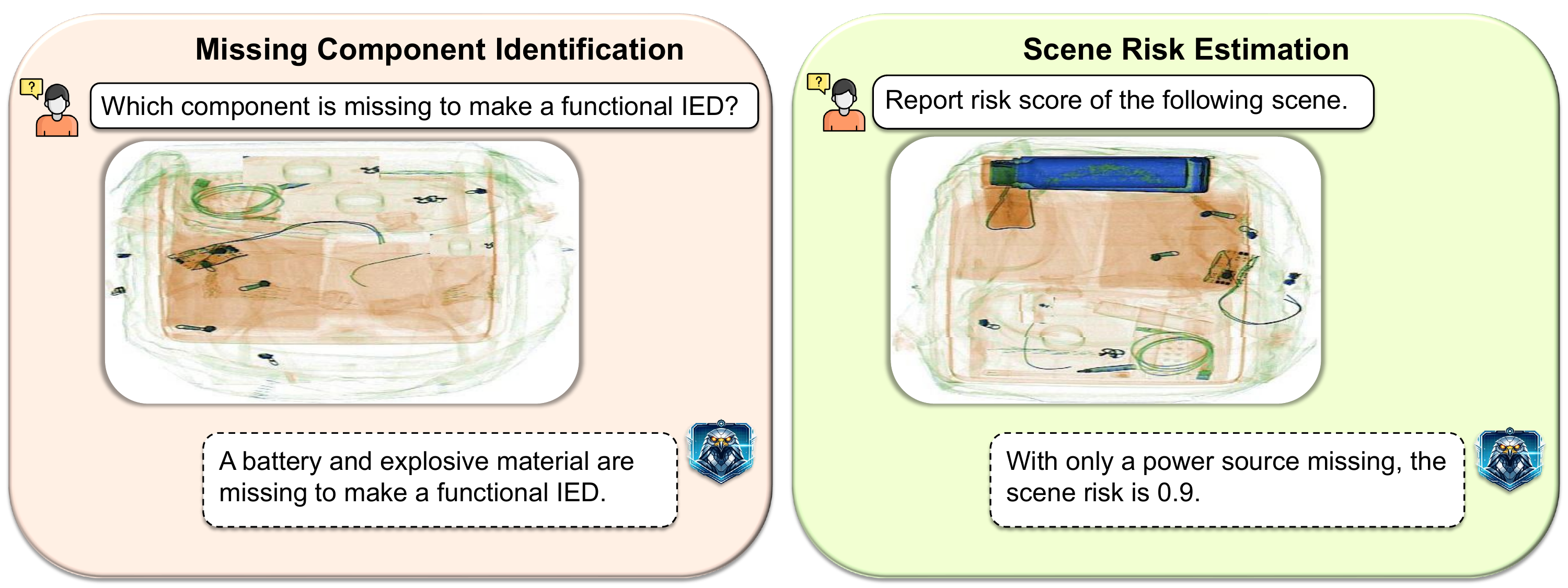}
\vspace{-1em}
\caption{\textbf{Falcon} enables segmentation-aware functional threat reasoning in X-ray imagery. It performs instance grounding, component presence recognition, referring functional grounding and more, while generating grounded and domain-aware natural-language explanations. Falcon reasons over spatially distant components of IED under heavy clutter, supporting structured threat analysis beyond object detection.}
\label{fig:intro}
\end{figure*}

\vspace{-3em}

\begin{abstract}

Conventional vision-language models are largely object-centric, focusing on detecting and describing individual entities. In safety-critical X-ray baggage screening, however, threat often emerges not from a single object but from the functional compatibility of spatially dispersed components, such as batteries, detonators, and explosive charges. We formalize this setting as \emph{compositional threat reasoning}, where risk is modeled as a relational property of grounded regions rather than an independent detection outcome. We introduce \textbf{Falcon}, a multimodal framework that abstracts segmentation-aware region features into a structured safety state capturing component presence, pairwise functional compatibility, and scene-level risk. This structured representation is injected into the language model as an explicit intermediate interface, encouraging relationally consistent and safety-aware reasoning. To evaluate this problem, we present \textbf{Falcon-X}, a benchmark that unifies dense grounding with structured supervision over component completeness and risk inference in cluttered X-ray imagery. Experiments show that while existing multimodal models adapt to appearance, they struggle with compositional safety reasoning. Falcon improves functional grounding and produces more coherent threat assessments, establishing compositional safety reasoning as a distinct evaluation paradigm for multimodal systems.

\end{abstract}
\vspace{-2em}

\section{Introduction}
\label{sec:intro}

\noindent X-ray baggage screening is a safety-critical task characterized by heavy clutter, material transparency, and severe object overlap \cite{survey}. Recent deep learning approaches have significantly advanced prohibited-item detection \cite{aodetr,lih,detectingphysical} on large-scale X-ray benchmarks \cite{sixray,pidray,hixray,clcxray,gdxray}, and multimodal models have enabled captioning and VQA in this domain \cite{stingbee}. However, existing systems remain fundamentally object-centric and threat is inferred from the presence of individual categories. In operational settings, however, risk is often \emph{compositional}. Improvised explosive devices (IEDs) may be transported in dismantled form, with batteries, detonators, and explosive charges spatially separated within a bag. Individually benign, these components become hazardous only through \emph{functional compatibility}. The central question is therefore not object recognition, but relational safety inference:
\begin{center}
\emph{Can spatially separated components under severe superposition collectively constitute a functional threat?}
\end{center}

\noindent Addressing this setting requires structured reasoning over multiple grounded regions and their interactions. Current X-ray benchmarks evaluate object-level detection, while multimodal large language models (MLLMs) \cite{blip2,llava,lisa,groma,glamm,kosmos2,shikra,gpt4roi} rely on implicit attention without explicit supervision over inter-component compatibility. Consequently, relationally consistent safety inference remains largely unexplored in cluttered X-ray imagery.

We formalize \emph{compositional threat reasoning} as a structured multimodal prediction problem in which safety is modeled as a relational property of grounded components. We introduce \textbf{Falcon}, a segmentation-aware multimodal framework that maps region-level features to an explicit safety state comprising component presence, pairwise functional compatibility, and calibrated scene-level risk. This structured state is injected into the language model via a \textbf{Structured Safety Adapter (SSA)}, enforcing relational consistency between perception and reasoning (Fig. \ref{fig:intro}).

\begin{table*}[!h]
\centering

\caption{
Comparison of Falcon-X with existing X-ray security benchmarks.
DC denotes explicit modeling of dismantled components.
Falcon-X is the only dataset that jointly supports dense grounding,
multimodal understanding, and structured functional threat reasoning.
}
\vspace{-1em}
\setlength{\tabcolsep}{2pt}
		\scalebox{0.6}[0.6]{

\begin{tabular}{lll|cccc|cccccccc|c}

\hline

\rowcolor{Gray} \textbf{Dataset}
& \textbf{\# Samples}
& \textbf{DC}
& \multicolumn{4}{c|}{\textbf{Annotations}}
& \multicolumn{8}{c|}{\textbf{Task Suite}}
& Avail.
\\

\cline{4-15} 

\rowcolor{Gray} 
&
&
& \textbf{Labels} & \textbf{Bbox} & \textbf{Mask} & \textbf{Caption} 
& \textbf{RefSeg} & \textbf{RefPan} & \textbf{Pan} & \textbf{VQA} & \textbf{RL} & \textbf{FG} & \textbf{GG} & \textbf{SA}
&
\\

\hline
 
GDXray (JNDE'15)\cite{dvxray} &  8,150 & \xmark & \cmark & \cmark & \xmark & \xmark & \xmark & \xmark & \xmark & \xmark & \xmark & \xmark & \xmark & \xmark & \cmark \\

SIXray (CVPR'19)\cite{sixray}  & 8,929 & \xmark & \cmark & \cmark & \xmark & \xmark & \xmark & \xmark & \xmark & \xmark & \xmark & \xmark & \xmark & \xmark & \cmark \\

OPIXray (ACMMM'20)\cite{opixray} & 46,642 & \xmark & \cmark & \cmark & \xmark & \xmark & \xmark & \xmark & \xmark & \xmark & \xmark & \xmark & \xmark & \xmark & \cmark \\

HiXray (ICCV'21)\cite{hixray}  & 45,364 & \xmark & \cmark & \cmark & \xmark & \xmark & \xmark & \xmark & \xmark & \xmark & \xmark & \xmark & \xmark & \xmark & \cmark \\

CLCxray (ITIFS'22)\cite{clcxray}& 9,565 & \xmark & \cmark & \cmark & \xmark & \xmark & \xmark & \xmark & \xmark & \xmark & \xmark & \xmark & \xmark & \xmark & \cmark \\

PIXray (ITMM'22)\cite{pixray}& 5,046 & \xmark & \cmark & \cmark & \cmark & \xmark & \xmark & \xmark & \xmark & \xmark & \xmark & \xmark & \xmark & \xmark & \cmark \\

EDS (CVPR'22)\cite{eds} & 14,219 & \xmark & \cmark & \cmark & \xmark & \cmark & \xmark & \xmark & \xmark & \xmark & \xmark & \xmark & \xmark & \xmark & \cmark \\

LPIXray (CIPAE'23)\cite{lpixray}& 60,950 & \xmark & \cmark & \cmark & \xmark & \xmark & \xmark & \xmark & \xmark & \xmark & \xmark & \xmark & \xmark & \xmark & \xmark \\

PIDray (IJCV'23)\cite{pidray}  & 47, 677 & \xmark & \cmark & \cmark & \cmark & \xmark & \xmark & \xmark & \xmark & \xmark & \xmark & \xmark & \xmark & \xmark & \cmark \\

DvXray (ITIFS'24)\cite{dvxray}& 5,496 & \xmark & \cmark & \cmark & \xmark & \xmark & \xmark & \xmark & \xmark & \xmark & \xmark & \xmark & \xmark & \xmark & \cmark \\

STCray (CVPR'25)\cite{stingbee} & 46,642 & \xmark & \cmark & \cmark & \cmark & \cmark & \xmark & \xmark & \xmark & \xmark & \xmark & \xmark & \xmark & \xmark & \cmark \\

\midrule

\rowcolor{gaincolor} \textbf{Falcon-X} & 
\textbf{6,911} & \cmark & \cmark & \cmark & \cmark & \cmark & \cmark & \cmark & \cmark & \cmark & \cmark & \cmark & \cmark & \cmark & \cmark \\ 

\hline

\end{tabular}
}
\label{tab:datasets}
\vspace{-1.2em}
\end{table*}

\begin{figure*}[h!]
\centering
\includegraphics[width=0.97\linewidth,height=6cm]{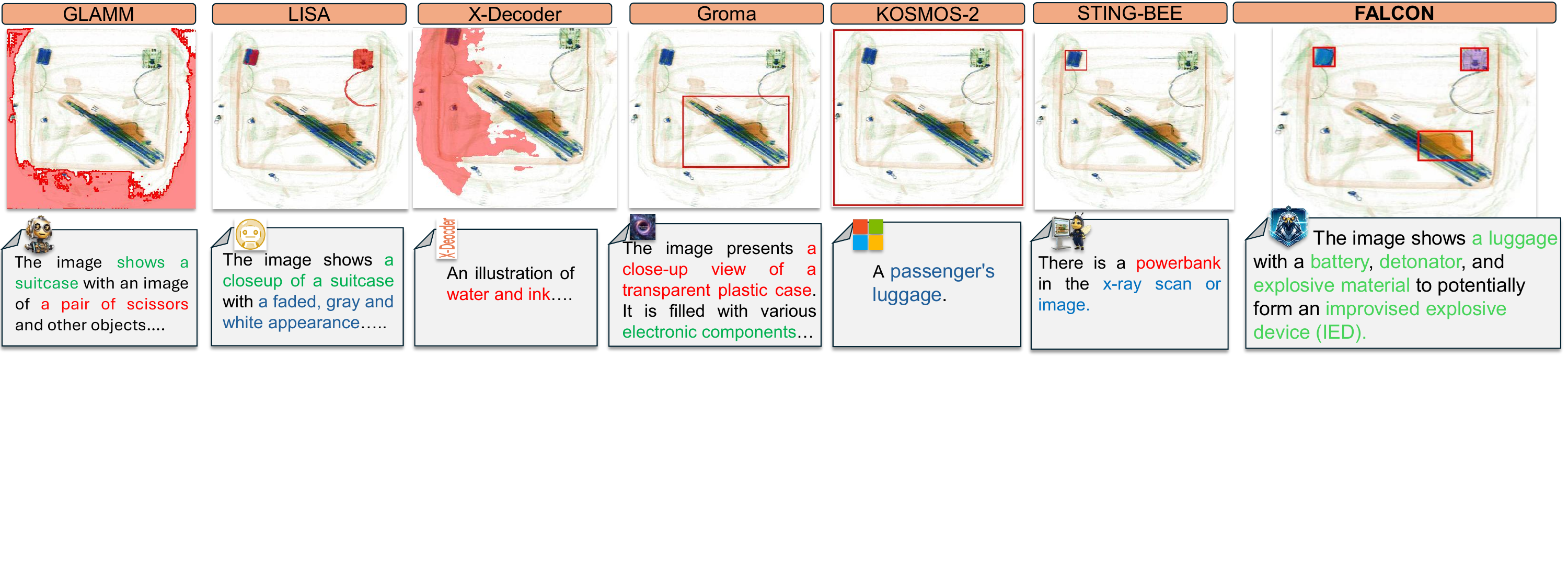}
\vspace{-8em}
\caption{
\textbf{Qualitative comparison of structured threat reasoning.} Existing models, including X-ray domain VLMs, do not model functional relation of components. \textcolor{red}{RED:} indicates hallucinations, \textcolor{blue}{BLUE:} indicates weak semantic relevance, and \textcolor{green}{GREEN:} indicates correct description.
}
\vspace{-2em}
\label{fig:qualitativefigure}

\end{figure*}
To evaluate this setting, we introduce \textbf{Falcon-X}, a benchmark of $\sim$7,000 real X-ray scans with instance-level annotations for key IED components. Beyond object detection, Falcon-X provides structured supervision for component completeness and calibrated risk, enabling evaluation of relational grounding and compositional safety reasoning under severe superposition. We benchmark Falcon-X across general-purpose and domain-adapted VLMs (Fig. \ref{fig:qualitativefigure}).

\noindent Our contributions are threefold:
\begin{itemize}
    \item We formalize compositional threat reasoning in X-ray imagery, modeling safety as a relational property of spatially grounded components.
    \item We introduce Falcon-X, the first benchmark coupling dense grounding with structured supervision for functional completeness and safety-consistent risk inference.
    \item We propose Falcon, a structured multimodal framework that injects an explicit safety state into region-level MLLMs to improve relationally coherent threat assessment.
\end{itemize}

\vspace{-1em}

\section{Related Work}
\noindent \textbf{General MLLMs:} Multimodal large language models (MLLMs) extend language models to vision by coupling visual encoders with language backbones. Early methods \cite{clip,blip2} demonstrate the effectiveness of large-scale image-text pretraining for transferrable multimodal representations. Instruction tuned extensions \cite{llava,minigpt4} build conversational interfaces on top of frozen pretrained vision encoders to answer complex queries about whole images. To enable explicit target localization, recent works \cite{kosmos2,shikra,groma} incorporate region-level representations by encoding bounding boxes as tokens and \cite{gpt4roi} achieves that by extracting region features via pooling. \cite{lisa,glamm} further extend language models to generate pixel-level masks for fine-grained localization. Despite these advances, such MLLMs have been evaluated mainly on everyday images and visually separable scenarios, and they do not target safety-critical tasks to explicitly reason over functional relationships or to detect disassembled threat components under heavy occlusion. \\

\noindent \textbf{X-ray baggage analysis:} X-ray security images pose unique challenges due to material transparency, object overlap and heavy clutter. Large-scale benchmarks \cite{gdxray,pidray,opixray,sixray,hixray} have enabled progress in detecting prohibited items under the above conditions. These datasets, predominantly follow a closed-set, object-centric paradigm, focusing on intact common threat categories such as guns, knives and scissors. Recent MLLM work introduced image-text paired dataset STCray \cite{stingbee} for extended tasks of image captioning, target localization and VQA in X-ray domain. However, existing benchmarks \cite{xssl} evaluate threats at the object level and do not explicitly model spatially scattered configurations of prohibited items in which risk arises from the functional composition of distributed components. Missing-part reasoning, functional sufficiency assessment and structured safety evaluation remain largely unexplored in this domain.

\vspace{-1em}

\section{Problem Setup: Compositional Threat Reasoning}
\label{sec:problem}

\noindent
We formalize dismantled IED assessment in X-ray imagery as a structured compositional inference problem. Unlike object-centric detection, where threat is attributed to isolated categories, we model safety as a relational property of spatially grounded components and their functional compatibility. Although Falcon-X instantiates this formulation with IED components, the same part-relation risk template can be extended to other modular threats by changing the component vocabulary and compatibility rules.

\vspace{-1em}
\subsection{Structured Safety State}

Let $I$ denote a single-view X-ray image and 
$
\mathcal{C} = \{\text{battery}, \text{detonator}, \text{main charge}\},
$
a predefined functional component taxonomy. We define a binary presence vector
$
\mathbf{y} \in \{0,1\}^{|\mathcal{C}|}, \quad
y_c = 1 \;\text{iff at least one instance of component } c \in \mathcal{C} \text{ is present}.$ To capture functional compatibility, we define a deterministic functional compatibility template $ \mathbf{L} \in [0,1]^{|\mathcal{C}| \times |\mathcal{C}|}, $ where $L_{uv}$ encodes the relative functional compatibility strength between component types $u$ and $v$. Higher values indicate stronger likelihood that the two components can jointly participate in a functional assembly. The matrix $\mathbf{L}$ is predefined at the type level and serves as a compatibility prior and annotation scaffold. The pair $(\mathbf{y}, \mathbf{L})$ defines a structured safety state.

\vspace{-1em}
\subsection{Compositional Threat Formulation}

Threat is treated as a relational property over present components. A minimal completeness condition is

\begin{equation}
\text{complete}(I) = \mathbb{1}\{\mathbf{y} = \mathbf{1}\},
\label{eq:complete}
\end{equation}

\noindent indicating that all required component types are observed under the predefined functional taxonomy. Functional completeness captures whether the required parts are visible under the type-level template. Beyond completeness, threat depends on the compatibility among the present components. Therefore, we define a continuous scene-level risk variable $
r \in [0,1],
$
representing the likelihood that grounded components admit a plausible functional assembly conditioned on $(\mathbf{y}, \mathbf{L})$.

\vspace{-1em}
\subsection{Learning Objective}

Given an image $I$ and optional query $q$, the model performs grounded perception, structured state estimation, and language generation. It first predicts segmentation-aware region proposals $R = \{(b_i, m_i, s_i)\}_{i=1}^{N},$ where $b_i$, $m_i$, and $s_i$ denote bounding box, mask, and confidence. From $R$, the model estimates $
\hat{\mathbf{p}} \in [0,1]^{|\mathcal{C}|}, \quad
\hat{\mathbf{L}} \in [0,1]^{|\mathcal{C}| \times |\mathcal{C}|}, \quad
\hat{r} \in [0,1],$ corresponding to component presence, image-conditioned functional links, and scene risk. These form a predicted safety state $(\hat{\mathbf{p}}, \hat{\mathbf{L}}, \hat{r})$. Finally, the model generates a response
$
Y = f(I,q),
$ grounded in the inferred structured state.

\vspace{-1em}
\section{Falcon-X Dataset}
\vspace{-0.5em}
\noindent We introduce \textbf{Falcon-X}, a dual-energy X-ray benchmark for compositional threat reasoning. Unlike existing datasets focused on intact prohibited-item detection, Falcon-X models spatially dispersed functional components whose risk arises from compatibility rather than object identity (Table \ref{tab:datasets}). The dataset combines dense instance grounding with structured supervision over component presence and scene-level risk, enabling evaluation of relational threat assessment.

\vspace{-1em}
\subsection{Data Collection}

Falcon-X contains 7,000 real dual-energy baggage scans with inert dismantled IED components drawn from $\mathcal{C}=\{\text{battery},\text{detonator},\text{main charge}\}$ \cite{eshetu2016inert}. Component combinations, spatial layouts, and occlusion levels are systematically varied to generate compositional configurations under realistic clutter and superposition. Additional details are provided in Appendix \ref{supplementary: Additional Details on Dataset collection}.

\vspace{-1em}

\begin{figure*}[h!]
\centering
\includegraphics[width=0.99\linewidth]{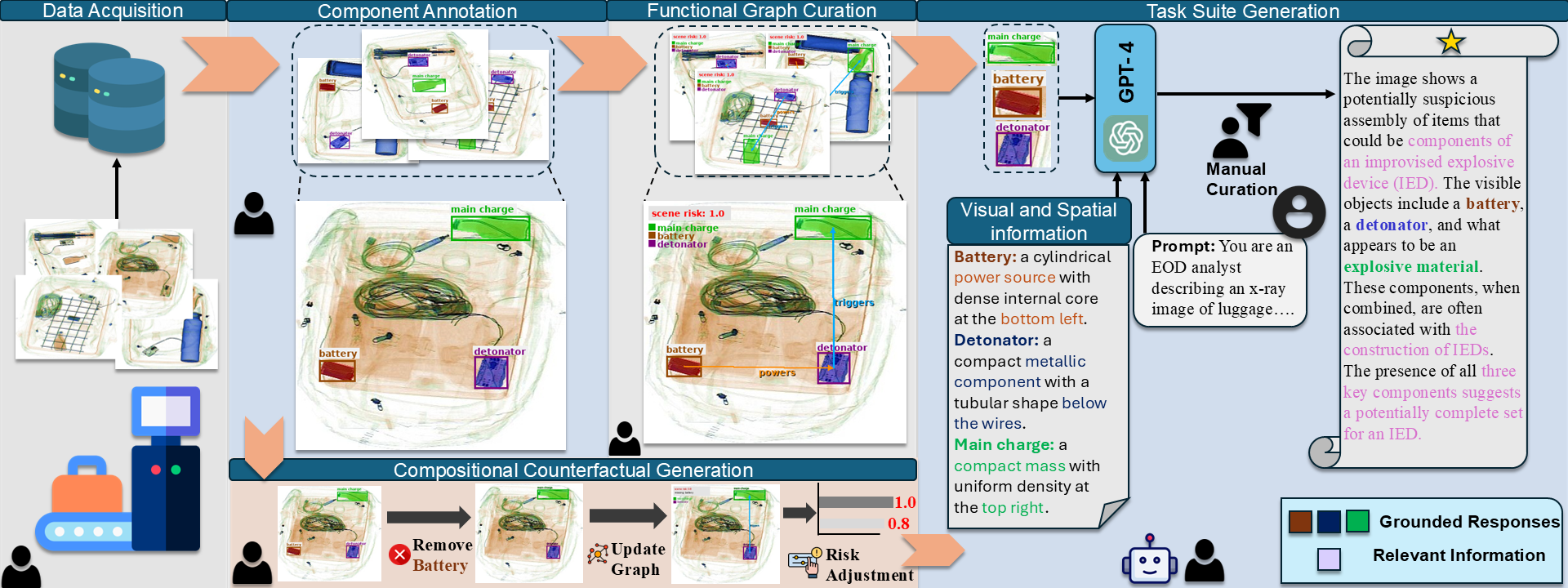}
\vspace{-0.5em}
\caption{
\textbf{Overview of the Falcon-X data collection pipeline.} Collected X-ray images undergo pixel-level component-wise annotations. \textbf{Functional Graph Curation} extends those annotations into pair-wise functionality links. \textbf{Compositional Counterfactual Generation} stage applies controlled counterfactual image generation with risk supervision. \textbf{Task Suite Generation} stage automates the process of Falcon-X task suite creation followed by manual curation. \includegraphics[height=1.3em]{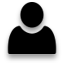}\hspace{0.1em} indicates human involvement, while \includegraphics[height=1.3em]{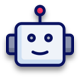}\hspace{0.1em} indicates automated stages.}
\label{fig:data_pipeline}
\vspace{-1.5em}
\end{figure*}

\vspace{-0.5em}
\subsection{Dense and Structured Annotation}
\label{subsec:annotation}

Each image is annotated with instance-level bounding boxes and pixel-accurate masks (Fig.\ref{fig:data_pipeline}). From these annotations, we derive the structured safety state comprising: (i) component presence $\mathbf{y}$, (ii) functional completeness (Eq. \ref{eq:complete}), and (iii) Link matrix $\mathbf{L}$ defined over $\mathcal{C}$. 

Each scene is additionally assigned a risk score $r \in [0,1]$ reflecting the likelihood that the observed component configuration constitutes a functional threat. Risk labels are provided by expert annotators and incorporate perceptual uncertainty: scenes missing one component may still receive high risk due to possible occlusion or concealment. This prevents trivial derivation of $r$ from $\mathbf{y}$ and enforces uncertainty-aware reasoning.

\vspace{-1em}
\subsection{Counterfactual Extension}

To evaluate missing-component and partial-assembly reasoning, we generate controlled counterfactual variants via mask-guided inpainting. Selected component instances are covered with a background mask obtained from the same image to preserve background clutter and semantic coherence, producing images with altered compositional states. This technique was chosen over mask zero-filling to avoid the introduction of visible artifacts. By enumerating feasible subsets of $\mathcal{C}$, we obtain a balanced corpus spanning single components, partial assemblies, and complete configurations. The final dataset comprises approximately 50,000 images (real + counterfactual). Details are provided in Appendix \ref{supplementarysubsec:synthetic}.

\vspace{-1em}
\subsection{Splits and Protocol}

Data are split at the base-image level (80/20 train/test), with all counterfactual variants assigned to the same partition to avoid leakage. Component distributions are preserved across splits. Annotations follow COCO format \cite{coco}, augmented with structured safety labels and multimodal task definitions (Fig.\ref{fig:data_pipeline}, Sec. \ref{sec:tasks}).


\vspace{-1em}
\section{Falcon-X Task Suite}
\label{sec:tasks}
\noindent
Falcon-X introduces a \emph{hierarchical evaluation framework} for compositional safety reasoning. 
The task suite progressively probes three levels of capability:  
(i) Grounded perception under X-ray superposition,  
(ii) compositional functional grounding, and  
(iii) Relationally consistent safety inference.

\noindent All tasks operate on an image $I$ and optional query $q$ or referring expression $t$. 
Segmentation outputs are evaluated against instance masks; language outputs are evaluated with standard captioning and QA metrics. 
Together, these tasks assess whether a model can construct, reason over, and coherently express a structured safety state (Fig. \ref{fig:tasksuite}). More details on task suite generation is in Appendix Sec.\ref{supplementary: Additional Details on FALCON-X task suite generation}.

\begin{figure*}[!t]
\centering
\includegraphics[width=0.99\linewidth]{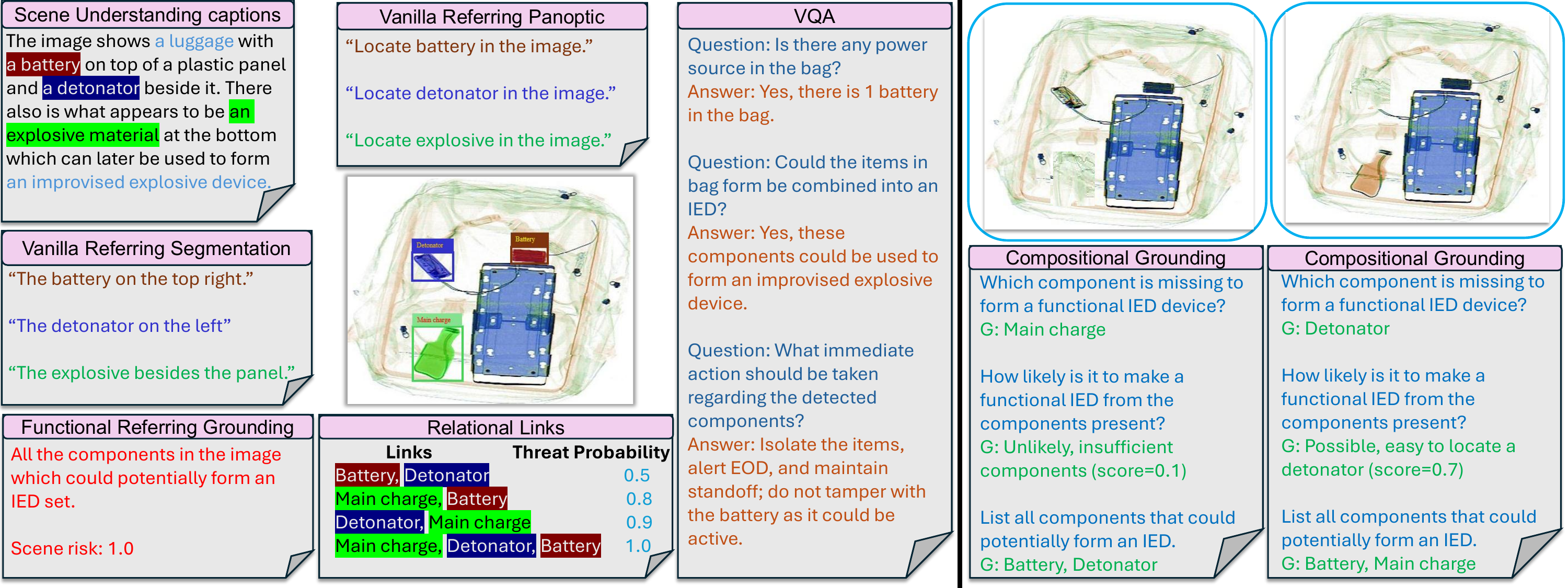}
\caption{
\textbf{Falcon-X Task Suite.} \textbf{Left:} Example of multi-level tasks on Falcon-X base image. \textbf{Right:} Samples of compositional grounding on controlled synthetic Falcon-X images for functional reasoning.}
\label{fig:tasksuite}
\vspace{-2em}
\end{figure*}

\subsection{Layer I: Grounded perception under X-ray superposition}
\label{subsec:layerI}
\noindent This layer evaluates whether a model can form an accurate perceptual state over component instances under heavy occlusion and transparency. 

\noindent \textbf{Scene understanding captions.} The model generates a scene description of $I$, that identifies and enumerates visible components in $\mathcal{C}$ and describes their spatial configuration. We report BLEU, METEOR, ROUGE-L, and CIDEr.

\noindent \textbf{Panoptic Segmentation.} Given $I$, the model predicts a panoptic map assigning each pixel a semantic label (battery/detonator/charge/background) and instance IDs. We report cIoU and mIoU.

\noindent \textbf{Referring Segmentation.} The model predicts masks from an image $I$, for the referred instance. Performance is measured with cIoU and mIoU.

\noindent \textbf{Referring Panoptic Segmentation.} From a given $t$, the model predicts a panoptic map of each pixel in $I$ belonging to the same semantic label while assigning the rest as background. Metrics of cIoU and mIoU are reported.

\noindent \textbf{Visual Question Answering.} We evaluate count and presence queries (e.g “How many batteries?”, “Is a detonator present?”). Metrics include exact-match accuracy and MAE for counting, and binary accuracy for presence.


\vspace{-1em}

\subsection{Layer II: Compositional Functional Reasoning}
\label{subsec:layerII}
\noindent This layer evaluates whether models can reason over sets of spatially dispersed components beyond conventional object detection.

\noindent \textbf{Missing Component Identification.} Given a complete or partial assembly of components defined in Sec.\ref{sec:problem} and the query $t$ “Which component is missing to form a functional IED?”, the model predicts the absent class from $\mathcal{C}$. We report performance as F1 and accuracy.

\noindent \textbf{Functional Completeness.} The model predicts a completeness score $\hat{c}\in[0,1]$ indicating whether the required component types are present and functionally compatible under the predefined component taxonomy. 
Functional completeness measures assembly sufficiency and does not directly encode expert uncertainty or contextual ambiguity. 
We report MAE and RMSE.

\noindent \textbf{Referring functional grounding.} Given a referring expression, "Ground all the components that could form a functional IED?", the model provides visually grounded regions of all possible components. Performance is measured in cIoU and mIoU.


\vspace{-1em}
\subsection{Layer III: Relationally consistent safety inference}
\label{subsec:layerIII}
\noindent This layer evaluates whether models produce calibrated, relationally coherent safety assessments over the full functional state $(\mathbf{y}, \mathcal{L}, r)$.

\noindent \textbf{Risk Prediction.} The model predicts a continuous scene-level risk score $\hat{r}\in[0,1]$. 
Risk is an expert-verified safety assessment conditioned on functional completeness, component compatibility, visual evidence, occlusion, and uncertainty. We report MAE.

\noindent \textbf{Functional Link Estimation.} The model predicts pairwise link probabilities $\hat{L}_{uv}$ indicating relational plausibility between component types. We evaluate using MAE for standard error assessment.

\noindent \textbf{Component Set Analysis.} We analyze logical consistency between predicted component presence $\hat{\mathbf{p}}$, link matrix $\mathcal{L}$, and risk $\hat{r}$ to query on grounding potential component sets. Incoherent predictions (e.g., high risk without functional completeness) are quantified to diagnose relational reasoning failures. We report set accuracy and F1 results.

\vspace{-1em}

\section{Falcon}
\vspace{-0.5em}
\noindent We propose \textbf{Falcon}, a structured segmentation-aware multimodal framework for compositional threat reasoning in X-ray imagery. Falcon introduces explicit semantic bottleneck between perception and language generation, enforcing structured safety inference prior to decoding:

\[
I \rightarrow R \rightarrow \{h_c\} \rightarrow (\hat{\mathbf{p}}, \mathcal{L}, \hat{r}) \rightarrow Y,
\]

\noindent where $R$ denotes instance regions, $\{h_c\}$ component-level slot embeddings, $\hat{\mathbf{p}}$ component presence, $\mathcal{L}$ functional links, $\hat{r}$ scene-level risk, and $Y$ the generated response. This intermediate state imposes a relational inductive bias, transforming threat assessment from implicit attention-based reasoning into structured inference.

\vspace{-1em}

\begin{figure*}[!t]
\centering
\includegraphics[width=0.99\linewidth,height=5cm]{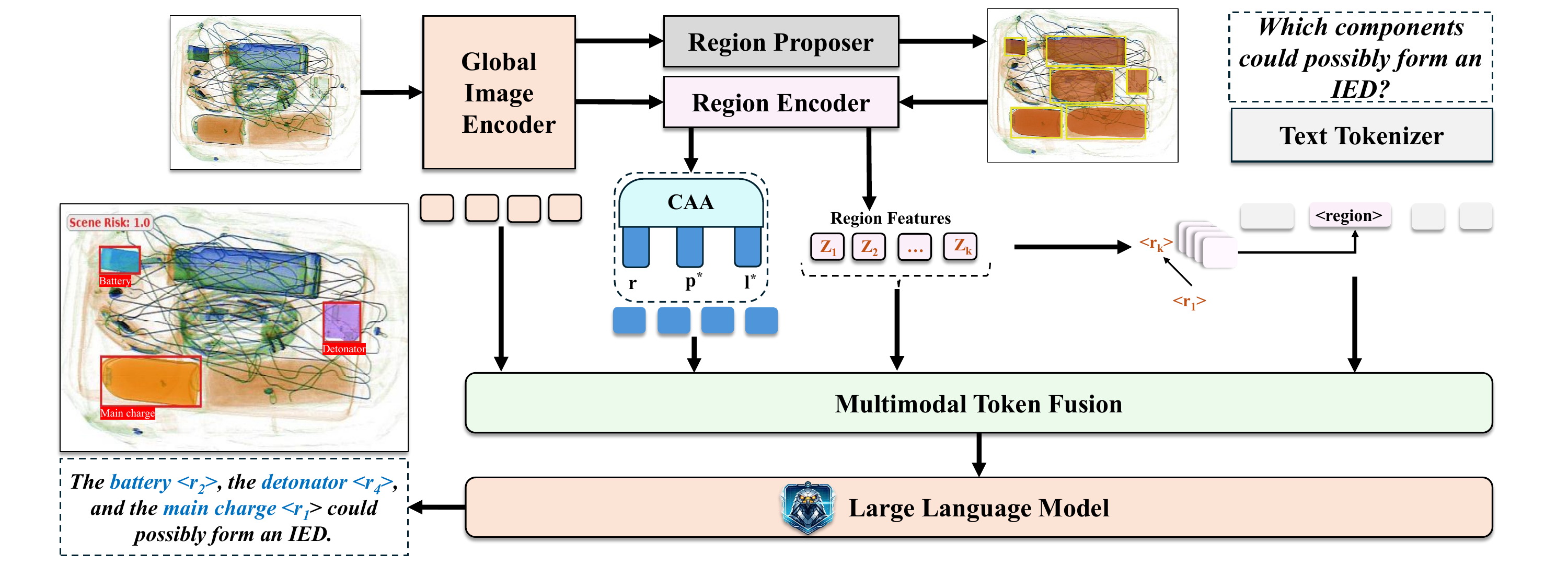}
\vspace{-1.3em}
\caption{\textbf{Falcon architecture.} Segmentation-aware perception extracts mask-aligned region embeddings, which are aggregated into structured component slots by the SSA to predict presence, functional links, and scene risk. The resulting structured tokens are fused with visual and textual tokens, introducing a relational perception before LLM decoding.}
\label{fig:block}
\vspace{-2.3em}
\end{figure*}

\subsection{Falcon Architecture}
\noindent Falcon integrates structured relational signals with region-level visual grounding to enable compositional reasoning over dismantled components.
The architecture consists of four parts:
(1) Segmentation-aware perception,
(2) Structured functional state modeling,
(3) Safety token injection, and 
(4) an LLM that performs multimodal reasoning and language generation. (Fig. \ref{fig:block})

\vspace{-1.5em}

\subsubsection{Segmentation-aware perception.}
Reliable functional reasoning in X-ray imagery benefits from instance-level grounding. This is achieved by three key components:


\noindent \textbf{Image encoder.}
We adopt DINOv2 \cite{dinov2} as the visual backbone and resize all X-ray inputs to $448 \times 448$. The encoder produces a dense grid of patch tokens that preserve spatial resolution and subtle contrast variations, which are critical for resolving small dismantled components under heavy clutter and translucency. 
\indent To reduce sequence length while maintaining spatial locality, non-overlapping $2 \times 2$ patches of tokens are linearly aggregated, following the token aggregation strategy adopted in \cite{groma}. This reduces the visual sequence by a factor of four while approximately preserving local structural cues. The resulting compact yet spatially structured feature map $F$ serves as the shared visual representation for downstream instance discovery and mask-aware grounding. \\

\noindent \textbf{Region proposer.}
A class-agnostic transformer detection head RF-DETR \cite{rfdetr} extracts instance proposals $R=\{(b_i,m_i,s_i)\}_{i=1}^N$, including bounding boxes, segmentation masks, and objectness confidence scores. Mask prediction is critical under X-ray superposition, where bounding boxes alone capture substantial background clutter. After confidence filtering and non-maximum suppression, the top-$K$ proposals are retained. The predicted instance masks provide precise spatial support for subsequent mask-aware region tokenization, enabling fine-grained grounding of partially occluded and fragmented components.

\noindent \textbf{Mask-aware region encoder.}
For each proposal in $R$, we combine ROI-aligned features $f_i^{\text{roi}}$, extracted from the backbone feature map $F$ using ROIAlign \cite{maskrcnn}, with mask-pooled features obtained from the predicted segmentation mask $m_i$ \(f_i^{\text{mask}} = \frac{\sum_{x} m_i(x)\,F(x)}{\sum_{x} m_i(x)}\), where $F(x)$ denotes the backbone feature at spatial location $x$, through a learnable projection. The fused region embedding \(z_i = W \,[f_i^{\text{roi}} \, \Vert \, f_i^{\text{mask}}]\) forms a language-aligned region token.

\vspace{-0.5em}
\subsubsection{Structured functional state modeling}
We introduce a Structured Safety Adapter (SSA) that maps a variable-sized set of region embeddings into a fixed, interpretable functional state representation.

\noindent \textbf{Query-based component slots (CAA).} Let $Z=\{z_i\}_{i=1}^{N},$ where $z_i \in \mathbb{R}^d,$ denote mask-aware region embeddings. SSA maintains $C=3$ learned component queries $Q=\{q_c\}_{c=1}^{3},$ where $q_c \in \mathbb{R}^d$, each corresponding to a predefined component type in Sec.\ref{sec:problem}. Each query attends to the region set via scaled dot-product attention as $A_{c,i} = 
\text{softmax}_i
\left(
\frac{q_c^\top z_i}{\sqrt d}
\right).
$ Component slot embeddings $h_c$ are obtained  from $Z$ by weighted aggregation as $h_c = \sum_{i=1}^{N} A_{c,i} z_i$ for $c = 1,2,3$. This query-based aggregation produces one structured embedding per component type, allowing multiple instances to contribute proportionally. Slot specialization emerges from component-level supervision during training. 

\noindent \textbf{Structured Predictions.} From $H=\{h_1,h_2,h_3\}$, SSA predicts three categories of structured signals:

\noindent \emph{(1) Component Presence.} For each component class $c \in \mathcal{C}$, a linear projection predicts its presence probability $\hat p_c = \sigma(f_p(h_c)).$

\noindent \emph{(2) Pairwise Functional Links.} For each predefined component pair $\mathcal{L}= \{(b,d),$ $(b,e),(d,e)\}$, we concatenate their corresponding embeddings and a lightweight MLP head predicts their pairwise functionality link to model a graph-like association between components $\hat \ell_{uv} = \sigma(f_\ell([h_u;h_v])), (u,v)\in\mathcal{P}.$ The link predictions model relational compatibility rather than physical connectivity.

\noindent \emph{(3) Scene Risk Level.} The scene-level risk is predicted from the structured component embeddings and inferred link probabilities, rather than directly from raw visual tokens. Specifically, a lightweight MLP head estimates a scalar value $\hat r =
\sigma\big(
f_r([h_1,h_2,h_3,\hat{\ell}_{bd},\hat{\ell}_{be},\hat{\ell}_{de}])
\big),$ $\hat{r}\in [0,1].$\\

\vspace{-0.7em}
\noindent \textbf{Tokenized Safety Conditioning.}
Structured predictions are assembled into
$
v = [\hat r,\hat p_1,\hat p_2,\hat p_3,\hat \ell_{bd},\hat \ell_{be},\hat \ell_{de}].
$ Each scalar $v_k$ is converted to a safety token
$t_k = e^{type}_k + \phi(v_k),$ where $e^{type}_k$ encodes variable identity and $\phi$ projects the value into the LLM embedding space. The resulting tokens are concatenated with image and region tokens prior to decoding, enabling conditioning on the inferred structured safety state.


\noindent \textbf{LLM.}
We use Vicuna-7B \cite{vicuna} as the language backbone. Image, region, and safety tokens are projected into the LLM embedding space and concatenated with text tokens to form
$[T_{\text{image}} \Vert T_{\text{region}} \Vert T_{\text{SSA}} \Vert T_{\text{text}}].$ The LLM is frozen and adapted using LoRA in attention layers for efficient domain-specific fine-tuning.

\vspace{-1.3em}

\subsection{Model Training}
\label{subsec:modeltraining}
Falcon is trained in three stages to decouple segmentation learning from structured multimodal reasoning. \\

\noindent \textbf{Stage 1: Segmentation-Aware Proposal Learning.}
We train RF-DETR \cite{rfdetr} in a class-agnostic manner to produce instance proposals and masks, treating all foreground components as a single threat class $\mathcal{L}_{S1} = \mathcal{L}_{det} + \mathcal{L}_{seg},$ where $\mathcal{L}_{det}$ denotes standard DETR set prediction losses (classification, box regression, GIoU), and $\mathcal{L}_{seg}$ combines mask BCE and Dice terms. Only RF-DETR is optimized; all multimodal modules remain frozen. The resulting proposals are fixed for subsequent stages. \\

\noindent \textbf{Stage 2: Structured Multimodal Alignment.}
With RF-DETR and the LLM frozen, region proposals are converted into mask-aware embeddings $Z$. The Structured Safety Adapter (SSA) maps $Z$ to component slots and predicts scene risk $\hat r$, presence $\hat{\mathbf p}$, and relational links $\hat{\mathbf L}$, which are injected as structured tokens into the LLM. The training objective jointly supervises language generation and structured safety prediction:

{\small
\begin{align}
\mathcal{L}_{S2}
= - \sum_t \log 
P_\theta(y_t \mid y_{<t}, I, Z, \text{SSA})
+ \lambda_r \|\hat r - r\|_1
+ \lambda_p \|\hat{\mathbf p} - \mathbf p\|_1
+ \lambda_\ell \|\hat{\mathbf L} - \mathbf L\|_1,
\end{align}
}

\noindent Here $P_\theta$ denotes the autoregressive LLM generation process. Only the SSA and projection layers are updated. \\

\noindent \textbf{Stage 3: Instruction Fine-Tuning.}
Starting from the Stage-2 checkpoint, we enable LoRA \cite{lora} adapters within the LLM while keeping RF-DETR frozen. The optimization objective remains $\mathcal{L}_{S2}$, but gradients now propagate through the LoRA parameters, improving instruction adherence and alignment between structured predictions and generated explanations.


\vspace{-1.5em}
\section{Experiments}
\vspace{-0.5em}
\subsection{Experimental Setup}

We use DINOv2-L/14 \cite{dinov2} as the visual backbone with $448\times448$ inputs. Region proposals are generated by class-agnostic RF-DETR (seg-2xlarge, 6 decoder layers) \cite{rfdetr}. Up to 300 proposals are produced, followed by NMS (0.6) and score filtering (0.15), retaining the top 100 regions. Mask-aware region features are projected to the LLM space and aggregated by the SSA to predict risk, component presence, and pairwise links, forming seven structured safety tokens. Training follows three stages: (i) 12 epochs of detector pretraining, (ii) 1 epoch of structured multimodal alignment with frozen visual backbone and LLM, and (iii) 1 epoch of instruction tuning using LoRA \cite{lora} (rank 16). Optimization uses AdamW with mixed precision on two A100 GPUs. Additional details are provided in the Appendix Sec.\ref{supplementary: Additional Details on Experiment Setup}.

\vspace{-1.2em}

\subsection{Results Discussion}
Following Sec.\ref{sec:tasks}, fine-tuned models are trained and evaluated on the Falcon-X train/test splits, while zero-shot methods are evaluated on the test split only. Top three results per metric are highlighted in \textcolor{red}{Red}/\textcolor{green}{Green}/\textcolor{blue}{Blue}, and Falcon's margin over the next best method is shown in \textcolor{gray}{Gray}. Cross-dataset generalization results are provided in Appendix \ref{supplementary: Cross-dataset evaluation}. \\
\vspace{-0.1em}
\noindent \textbf{Scene Understanding and VQA.}
Table \ref{tab:su} evaluates scene understanding captioning and VQA performance of models on Falcon-X (Sec.\ref{subsec:layerI}). Zero-shot general-purpose vision-language models exhibit limited transfer to X-ray imagery, reflecting the substantial distribution gap between natural images and heavily overlapped security scans. Moreover, this was also the case for domain-adapted methods like Sting-Bee \cite{stingbee}. After fine-tuning, however, performance saturates across all models, indicating that conventional scene understanding in this case is largely a matter of domain adaptation. Falcon performs on par with or slightly exceeding strong baselines such as Groma \cite{gres}, LISA \cite{lisa} and Sting-Bee \cite{stingbee}. Notably, the integration of structured functional state modeling preserves general multimodal capabilities while targeting higher-level compositional inference.
\vspace{-1.7em}

\begin{table*}[!h]
\centering
\caption{Scene understanding image captioning and VQA performance on Falcon-X.}
\vspace{-1em}
\setlength{\tabcolsep}{2pt}
		\scalebox{0.7}[0.7]{

\begin{tabular}{l|ccc|cccc}

\hline

\rowcolor{Gray} 
 Method
& \multicolumn{3}{c|}{Image Captioning}
& \multicolumn{4}{c}{VQA}
\\

\cline{2-8}

\rowcolor{Gray} 

& BLEU  & ROUGE-L & CIDEr 
& BLEU & METEOR & ROUGE-L & CIDEr 
\\

\hline

\hline

 \textit{\textbf{Zero-shot}} 
&  &    & 
&  &  &  & 
\\

 Sa2Va \cite{sa2va} 
& 25.06 &  18.11 & 0.012
& 2.88 & - & 3.25  & -
\\

 Kosmos-2 \cite{kosmos2} 
& 16.26 &  17.56 & 0.002
& 2.83 & 5.47 & 3.79 & 0.003
\\

 Llava-1.5 \cite{llava}
 & 12.67 &  15.67 & -
& 2.77 & 7.64 & 3.39 & -
\\

 GLaMM \cite{glamm}
& 15.91 &  17.56 & 0.0032
& 1.95 & 4.86 & 3.52 & 0.001
\\

 LISA \cite{lisa}
& 20.57 &  18.08 & 0.01
& 3.86 & - & 3.94 & -
\\

 Groma \cite{groma}
& 18.8 &  16.48 & 0.004
& 3.32 & 8.90 & 3.44 & - 
\\

 Sting-Bee \cite{stingbee}
 & 24.61 &  17.95 & 0.012
& 3.86 & 7.74 & 4.04 & 0.01
\\

\cline{1-8}

 \textit{\textbf{Fine-tuned}} 
&  &  &  
&  &  &  & 
\\

 LISA \cite{lisa}
& \textcolor{blue}{35.57} & \textcolor{blue}{39.13} & \textcolor{blue}{0.04}
& \textcolor{green}{99.91} & \textcolor{green}{91.42} & \textcolor{green}{99.93} & \textcolor{green}{6.25}
\\

Sting-Bee \cite{stingbee}
& 33.38 & 37.48 &  0.024
& 99.12 & \textcolor{blue}{91.27} & \textcolor{blue}{99.92} & \textcolor{blue}{6.24}
\\

Groma \cite{groma} 
& \textcolor{green}{40.7} & \textcolor{green}{47.69} & \textcolor{green}{0.061}
& \textcolor{blue}{99.86} & 91.24 & \textcolor{blue}{99.92} & \textcolor{blue}{6.24}
\\

\rowcolor{gaincolor}  Ours 
& 
\textcolor{red}{40.92} \footnotesize{(\textcolor{gray}{\textbf{0.22$\uparrow$}})}
& \textcolor{red}{47.74} \footnotesize{(\textcolor{gray}{\textbf{0.05$\uparrow$}})} & \textcolor{red}{0.064} \footnotesize{(\textcolor{gray}{\textbf{0.003$\uparrow$}})}
& \textcolor{red}{99.93} \footnotesize{(\textcolor{gray}{\textbf{0.02$\uparrow$}})} & \textcolor{red}{91.98} \footnotesize{(\textcolor{gray}{\textbf{0.56$\uparrow$}})} & \textcolor{red}{99.95} \footnotesize{(\textcolor{gray}{\textbf{0.02$\uparrow$}})} & \textcolor{red}{6.26} \footnotesize{(\textcolor{gray}{\textbf{0.01$\uparrow$}})}
\\ 

\hline

\end{tabular}
}
\label{tab:su}
\vspace{-1.5em}
\end{table*}

\noindent \textbf{Perception and Compositional Grounding.}
Table \ref{tab:cfg} addresses the grounding tasks of Layers I and II evaluation in Sec.\ref{subsec:layerI} and \ref{subsec:layerII}. On conventional localization tasks (RS, PS, RPS), strong zero-shot models such as Sa2Va \cite{sa2va} remain competitive, and fine-tuned baselines achieve solid performance through domain adaptation. We do not consistently outperform prior methods on these appearance-driven tasks. In contrast, a distinct pattern emerges for referring functional grounding (RFG), where the target region is defined by functional composition rather than object identity. On RFG, our model outperforms the next strongest baselines by +30.38 and +40.03 points in cIoU and mIoU respectively. Notably, this relative gain is substantially larger than any difference observed on RS or RPS, indicating that the improvement does not arise from generic localization advances. The RFG improvement confirms that relational supervision enables compatibility-aware grounding. When targets are compositional rather than appearance-driven, structured modeling is essential. \\

\begin{table*}[!h]
\centering
\caption{Perception and compositional grounding performance on Falcon-X. Referring segmentation, Panoptic segmentation, Referring Panoptic Segmentation, and Referring Functional Grounding are represented as RS, PS, RPS and RFG, respectively.}

\vspace{-1em}

\setlength{\tabcolsep}{2pt}
		\scalebox{0.6}[0.6]{

\begin{tabular}{l|cc|cc|cc|cc}

\hline

\rowcolor{Gray} 
 Method
& \multicolumn{2}{c|}{RS}
& \multicolumn{2}{c|}{PS}
& \multicolumn{2}{c|}{RPS}
& \multicolumn{2}{c}{RFG}
\\

\cline{2-9}

\rowcolor{Gray} 

& cIoU &  mIoU
& cIoU & mIoU
& cIoU & mIoU
& cIoU &  mIoU
\\

\hline

 \textit{\textbf{Zero-shot}} 
&  &  
&  &  
&  &  
&  &  
\\

 Sa2Va \cite{sa2va} 
& 14.96 &  \textcolor{green}{32.78}
& \textcolor{red}{35.53} & \textcolor{red}{54.14}
& 17.5 & \textcolor{red}{37.98}
& \textcolor{blue}{14.26} & 20.64
\\

 Kosmos-2 \cite{kosmos2} 
& 2.95 & 3.2
& 2.78 & 3.08
& 2.92 & 3.32
& 3.61 & 3.62
\\

 GLaMM \cite{glamm} 
& 12.37 & 19.84 
& 6.41 & 13.72 
& 11.28 & 18.93 
& 5.73 & 17.02 
\\

 LISA \cite{lisa}
& 3.50 &  6.73 
& 9.13 &  14.21 
& 11.50 &  16.95 
& 11.86 &  13.24 
\\

 Groma \cite{groma}
 & 11.61 &  20.49

& 7.88 &  16.04 
& 14.06 &  21.2 
& 8.16 &  \textcolor{blue}{26.31} 
\\

 Sting-Bee \cite{stingbee}
& \textcolor{blue}{15.23} & 22.37   
& 12.45 & 15.69   
&13.21 & 14.95   
& 10.45 & 11.89   
\\

\cline{1-9}

 \textit{\textbf{Fine-tuned}} 
&  &  
&  &  
&  &  
&  &  
\\

 LISA \cite{lisa}
& 11.82 &  14.75
& \textcolor{blue}{35.18} &  \textcolor{blue}{37.12}
& \textcolor{red}{30.12} &  \textcolor{blue}{33.34}
& 9.45 & 12.98
\\

 Sting-Bee \cite{stingbee}
& \textcolor{green}{18.72} & \textcolor{blue}{26.81}
& 31.05 &  33.49
& 15.40 &  19.59
& 12.72 &  14.96
\\

 Groma \cite{groma}
& 14.25 &  21.99 
& \textcolor{green}{36.73} &  13.78
& \textcolor{blue}{17.94} &  24.1
& \textcolor{green}{20.07} &  \textcolor{green}{29.55}
\\

\rowcolor{gaincolor}  Ours 
& \textcolor{red}{25.86} \footnotesize{(\textcolor{gray}{\textbf{7.14$\uparrow$}})}  &  \textcolor{red}{35.39}
\footnotesize{(\textcolor{gray}{\textbf{2.61$\uparrow$}})}
& 14.37 \footnotesize{(\textcolor{gray}{\textbf{21.16$\downarrow$}})}
&  \textcolor{green}{38.02}
\footnotesize{(\textcolor{gray}{\textbf{16.12$\downarrow$}})}
& \textcolor{green}{21.74} \footnotesize{(\textcolor{gray}{\textbf{8.38$\downarrow$}})}
&  \textcolor{green}{35.30}\footnotesize{(\textcolor{gray}{\textbf{2.68$\downarrow$}})}

& \textcolor{red}{50.45} \footnotesize{(\textcolor{gray}{\textbf{30.38$\uparrow$}})}
& \textcolor{red}{69.58} \footnotesize{(\textcolor{gray}{\textbf{40.03$\uparrow$}})}
\\ 

\hline

\end{tabular}
}
\label{tab:cfg}
\vspace{-1em}
\end{table*}

\noindent \textbf{Compositional Threat Reasoning.}
Table \ref{tab:ftr} reports semantic grounding (Layer II) and relational safety metrics (Layer III). Zero-shot MLLMs perform poorly on compositional tasks, particularly MCI and relational safety measures, indicating limited transfer from object-level semantics to functional reasoning. After fine-tuning, component presence (CPC) saturates across models ($\approx$98\%), suggesting that independent component recognition is not the main challenge. In contrast, relational metrics remain substantially harder. Accurate estimation of functional completeness (FC), scene risk (SRL), and link risk (CLR) requires compatibility-aware reasoning rather than isolated detection. Falcon achieves competitive CPC while consistently reducing MAE on FC, SRL, and especially CLR (0.005), indicating improved relational calibration. The gains concentrate on compatibility-dependent metrics despite near-saturated presence scores, demonstrating that improvements arise from structured modeling of component interactions rather than improved object classification.
\vspace{-1.5em}

\begin{table*}[!h]
\centering

\caption{Compositional Semantic Grounding and Safety Relation analysis performance on Falcon-X. Component Presence Check, Missing Component Identification, Functional Completeness, Scene Risk Level, Potential Component Sets, and Component Link Risk are represented as CPC, MCI, FC, SRL, PCS and CLR, respectively.}

\vspace{-1em}

\setlength{\tabcolsep}{2pt}
		\scalebox{0.55}[0.6]{

\begin{tabular}{l|cccccc|cccc}

\hline

\rowcolor{Gray} 
 Method
& \multicolumn{6}{c|}{Semantic Grounding}
& \multicolumn{4}{c}{Safety Relationship}
\\

\cline{2-11}

\rowcolor{Gray} 

& \multicolumn{2}{c|}{CPC}
& \multicolumn{2}{c|}{MCI}
& \multicolumn{2}{c|}{FC}
& \multicolumn{1}{c|}{SRL}
& \multicolumn{2}{c|}{PCS}
& \multicolumn{1}{c}{CLR}
\\

\cline{2-11}

\rowcolor{Gray} 

& Acc$\uparrow$ & F1$\uparrow$ 
& Acc$\uparrow$ & F1$\uparrow$
& MAE$\downarrow$ & RMSE$\downarrow$
& MAE$\downarrow$
& Acc$\uparrow$ & F1$\uparrow$
& MAE$\downarrow$
\\

\hline

 \textit{\textbf{Zero-shot}} 
&  &
&  &
&  &
&  &
&  &
\\

Sa2Va \cite{sa2va}
& \textcolor{blue}{58.4} & 46.2
& 12.6 & 15.8
& 0.60 & --
& 0.40 & 13.56
& 54.9 & --
\\

 Kosmos-2 \cite{kosmos2} 
& 57.09 & 36.36
& 14.16 & 15.26
& 0.57 & 0.67
& 0.43 & 14.3
& \textcolor{green}{57.1} & --
\\

 Llava-1.5 \cite{llava}
& 57.4 & 47.12
& 14.32 & 14.93
& 0.7 & --
& -- & 14.1
& 56.47 & --
\\

 GLaMM \cite{glamm}
& 55.8 & 35.9 
& 11.6 & 14.2 
& 0.52 & -- 
& 0.31 & 12.6 
& 56.2 & -- 
\\

 LISA \cite{lisa}
& -- & --
& 14.29 & 14.29
& 0.9 & 0.9
& -- & --
& -- & --
\\

 Groma \cite{groma}
& 57.1 & \textcolor{blue}{36.4}
& 13.3 & 19.9
& 0.59 & 0.69
& 0.36 & 14.63
& 57.05 & --
\\

 Sting-Bee \cite{stingbee}
& 57.14 & 36.36
& 14.29 & 14.29
& 0.49 & 0.48
& 0.42 & 0.01
& 42.86 & --
\\

\cline{1-11}

 \textit{\textbf{Fine-tuned}} 
&  &
&  &
&  &
&  &
&  &
\\

 LISA \cite{lisa} 
& - & -
& 17.63 & 19.25
& \textcolor{blue}{0.28} & \textcolor{blue}{0.35}
& \textcolor{green}{0.20} & 0.41
& 43.06 & \textcolor{green}{0.20} 
\\

Sting-Bee \cite{stingbee} 
& \textcolor{green}{98} & \textcolor{green}{98}
& \textcolor{blue}{93.45} & \textcolor{blue}{96.31}
& 0.031 & 0.36
& \textcolor{blue}{0.23} & \textcolor{blue}{14.89}
& 56.34 & \textcolor{blue}{0.24}
\\

 Groma \cite{groma} 
& \textcolor{green}{98} & \textcolor{red}{98}
& \textcolor{red}{97.1} & \textcolor{red}{98.6}
& \textcolor{green}{0.019} & \textcolor{green}{0.13}
& - & \textcolor{green}{14.9}
& \textcolor{blue}{57.06} & -
\\

\rowcolor{gaincolor}  Ours 
& \textcolor{red}{98.1} \footnotesize{(\textcolor{gray}{\textbf{0.1$\uparrow$}})}
& \textcolor{green}{97.97} \footnotesize{(\textcolor{gray}{\textbf{0.03$\downarrow$}})}
& \textcolor{green}{94.75} \footnotesize{(\textcolor{gray}{\textbf{2.35$\downarrow$}})}
& \textcolor{green}{97.33} \footnotesize{(\textcolor{gray}{\textbf{1.27$\downarrow$}})}
& \textcolor{red}{0.017} \footnotesize{(\textcolor{gray}{\textbf{0.002$\downarrow$}})}
& \textcolor{red}{0.09} \footnotesize{(\textcolor{gray}{\textbf{0.04$\downarrow$}})}
& \textcolor{red}{0.02} \footnotesize{(\textcolor{gray}{\textbf{0.18$\downarrow$}})}
& \textcolor{red}{15.1} \footnotesize{(\textcolor{gray}{\textbf{0.2$\uparrow$}})}
& \textcolor{red}{57.69} \footnotesize{(\textcolor{gray}{\textbf{0.59$\uparrow$}})}
& \textcolor{red}{0.005} \footnotesize{(\textcolor{gray}{\textbf{0.195$\downarrow$}})}
\\

\hline

\end{tabular}
}
\label{tab:ftr}
\vspace{-2.7em}
\end{table*}

\vspace{-0.7em}
\subsection{Qualitative Results}
\begin{figure*}[!h]
\vspace{-1.5em}
\centering
\includegraphics[width=0.95\linewidth,height=5cm]{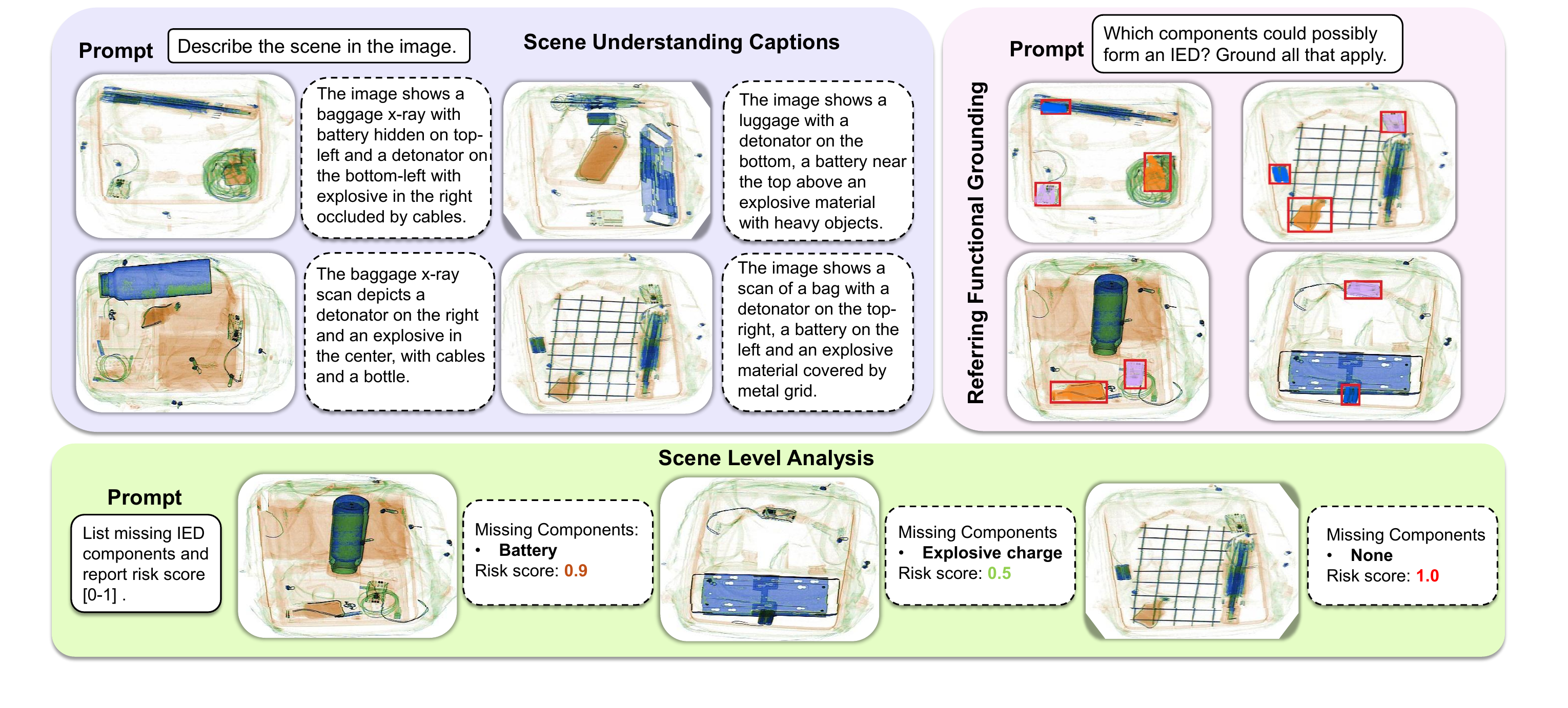}
\vspace{-1.5em}
\caption{Falcon qualitative results on compositional grounding and scene-level reasoning.}
\label{fig:qualitative}
\vspace{-2em}
\end{figure*}

Fig. \ref{fig:qualitative} shows qualitative examples of referring functional grounding and scene-level reasoning. Falcon localizes components that jointly satisfy functional assembly criteria, including spatially separated and partially occluded instances. For incomplete assemblies, it identifies missing components and adjusts risk predictions consistently with the inferred structured state. Additional examples are provided in Appendix \ref{supplementary: Additional Qualitative results}.

\vspace{-1.5em}
\subsection{Ablations}
We conduct controlled ablations to quantify the contribution of (i) structured prediction heads and (ii) perception versus reasoning bottlenecks. All variants are evaluated on the primary task of \textit{Referring Functional Grounding} (RFG). Additional ablations are reported in Appendix \ref{supplementary:Additional Ablation study}. \\

\noindent\textbf{Prediction head ablations.}
We ablate the Structured Safety Adapter (SSA) by selectively disabling the risk, presence, and link heads prior to token injection. As shown in Table \ref{tab:ablationheads}, removing all heads reduces Falcon to a region-grounded VLM without structured signals. Enabling the risk head alone yields limited improvement, while adding presence prediction provides a larger gain. The most significant increase occurs when the link head is activated, indicating that relational compatibility modeling is the primary driver of functional grounding performance.

\noindent\textbf{Oracle ablations.} Introduced to decouple the relative contributions of perception and structured reasoning modules (Table \ref{tab:ablationoracle}). \textit{Oracle-Reasoning} replaces the predicted functional selection with ground-truth relational assignments over the same predicted regions, isolating the decision layer under fixed perception. 
\textit{Oracle-Perception} replaces predicted regions with ground-truth masks while retaining learned reasoning, providing a perception upper bound. The gap Falcon$\rightarrow$Oracle-reasoning measures reasoning headroom, while Oracle-reasoning$\rightarrow$Oracle-perception quantifies the remaining perception bottleneck.

\vspace{-1.5em}

\begin{table*}[!h]
\begin{minipage}[t]{0.5\linewidth}
\centering
\caption{SSA Prediction heads ablations on referring functional grounding (RFG). R, P, and L denote the risk, component-presence, and link prediction heads. Performance is reported in cIoU and mIoU.}

\vspace{-1em}

\setlength{\tabcolsep}{2pt}
		\scalebox{0.8}[0.8]{

\begin{tabular}{ccc|cc}

\hline

\rowcolor{Gray} 

\multicolumn{3}{c|}{Variant}
& \multicolumn{2}{c}{RFG}
\\

\rowcolor{Gray} 
\multicolumn{1}{c}{R} & \multicolumn{1}{c}{P} &
\multicolumn{1}{c|}{L}
& cIoU & mIoU \\

\hline

\xmark & \xmark & \xmark & 35.82 & 43.43 \\

\cmark & \xmark & \xmark 

& 39.33 \footnotesize{(\textcolor{gray}{\textbf{3.51$\uparrow$}})}
& 47.98 \footnotesize{(\textcolor{gray}{\textbf{4.55$\uparrow$}})} \\

\cmark & \cmark & \xmark 
& 45.52 \footnotesize{(\textcolor{gray}{\textbf{6.19$\uparrow$}})}
& 58.60 \footnotesize{(\textcolor{gray}{\textbf{10.62$\uparrow$}})} \\

\rowcolor{gaincolor} \cmark & \cmark & \cmark 
& \textbf{50.45} \footnotesize{(\textcolor{gray}{\textbf{4.93$\uparrow$}})}
& \textbf{69.58} \footnotesize{(\textcolor{gray}{\textbf{10.98$\uparrow$}})} \\

\hline

\end{tabular}
\label{tab:ablationheads}
}
\end{minipage}
\hfill
\begin{minipage}[t]{0.44\linewidth}
\centering

\caption{Oracle ablation performance of Falcon perception and reasoning stages on referring functional grounding (RFG). Results are reported in cIoU and mIoU and $\Delta$vs. Falcon}

\vspace{-1em}

\setlength{\tabcolsep}{2pt}
		\scalebox{0.5}[0.8]{

\begin{tabular}{l|cc|cc}

\hline

\rowcolor{Gray} 
Mode &
\multicolumn{1}{c|}{Perception}
&\multicolumn{1}{c|}{Reasoning}
& \multicolumn{2}{c}{RFG} \\

\cline{4-5}

\rowcolor{Gray} 
& \multicolumn{1}{c|}{} &
 \multicolumn{1}{c|}{}
 & cIoU & mIoU \\

\hline

\rowcolor{gaincolor} \textbf{Falcon} & Pred & Pred
& 50.45  & 69.58   \\

Oracle-Reasoning & Pred & GT
& 59.61 $\Delta$\footnotesize{(\textcolor{gray}{\textbf{9.16$\uparrow$}})}
& 79.53 $\Delta$\footnotesize{(\textcolor{gray}{\textbf{9.95$\uparrow$}})} \\

Oracle-Perception & GT & Pred
& 79.49 $\Delta$\footnotesize{(\textcolor{gray}{\textbf{29.04$\uparrow$}})}
& 83.42 $\Delta$\footnotesize{(\textcolor{gray}{\textbf{14.84$\uparrow$}})} \\

\hline

\end{tabular}
\label{tab:ablationoracle}
}
\end{minipage}

\vspace{-3.5em}

\end{table*}

\section{Conclusion}
\vspace{-1em}
We formulate functional threat reasoning in X-ray imagery as a compositional inference problem, where safety emerges from the structured interaction of spatially dispersed components rather than isolated object detection. To address this setting, we introduce Falcon, a segmentation-aware multimodal framework that injects explicit structured relational state between perception and language generation, and Falcon-X, a benchmark that jointly evaluates grounding, compositional reasoning, and calibrated risk assessment. Through controlled experiments and diagnostic ablations, we show that explicit relational modeling substantially improves functional grounding and safety-consistent prediction under heavy superposition. By coupling structured supervision with multimodal reasoning, Falcon establishes a principled approach to compositional safety analysis, while Falcon-X provides a unified evaluation protocol for this emerging problem. We hope this work motivates further research on structured, safety-critical reasoning in complex visual environments.

\section*{Acknowledgements}
{\sloppy
This research was supported by Khalifa University Digital Future Institute (KU-DF), and the Khalifa University Center for Autonomous Robotic Systems (KU-CARS).}

\bibliographystyle{splncs04}
\bibliography{main}

\clearpage

\section*{Appendix}
\setcounter{page}{1}

\begin{itemize}
    \item Additional Details on Falcon-X (\autoref{supplementary: Additional Details on Dataset collection})
    \item Additional Details on Falcon-X Task Suite Generation (\autoref{supplementary: Additional Details on FALCON-X task suite generation})
    \item Qualitative Results (\autoref{supplementary: Additional Qualitative results})
    \item Cross-dataset Evaluation (\autoref{supplementary: Cross-dataset evaluation})
    
    \item Additional Details on Experimental Setup (\autoref{supplementary: Additional Details on Experiment Setup})
    \item Additional Ablation study (\autoref{supplementary:Additional Ablation study})
    \item Future Work (\autoref{supplementary: Future Work}) 
\end{itemize}

\section{Additional Details on Falcon-X}
\label{supplementary: Additional Details on Dataset collection}

\subsection{Acquisition Setup and Collection Protocol}
\label{supplementarysubsec:acquisition}
\begin{figure*}[!h]
\centering
\includegraphics[width=0.8\linewidth,height=5cm]{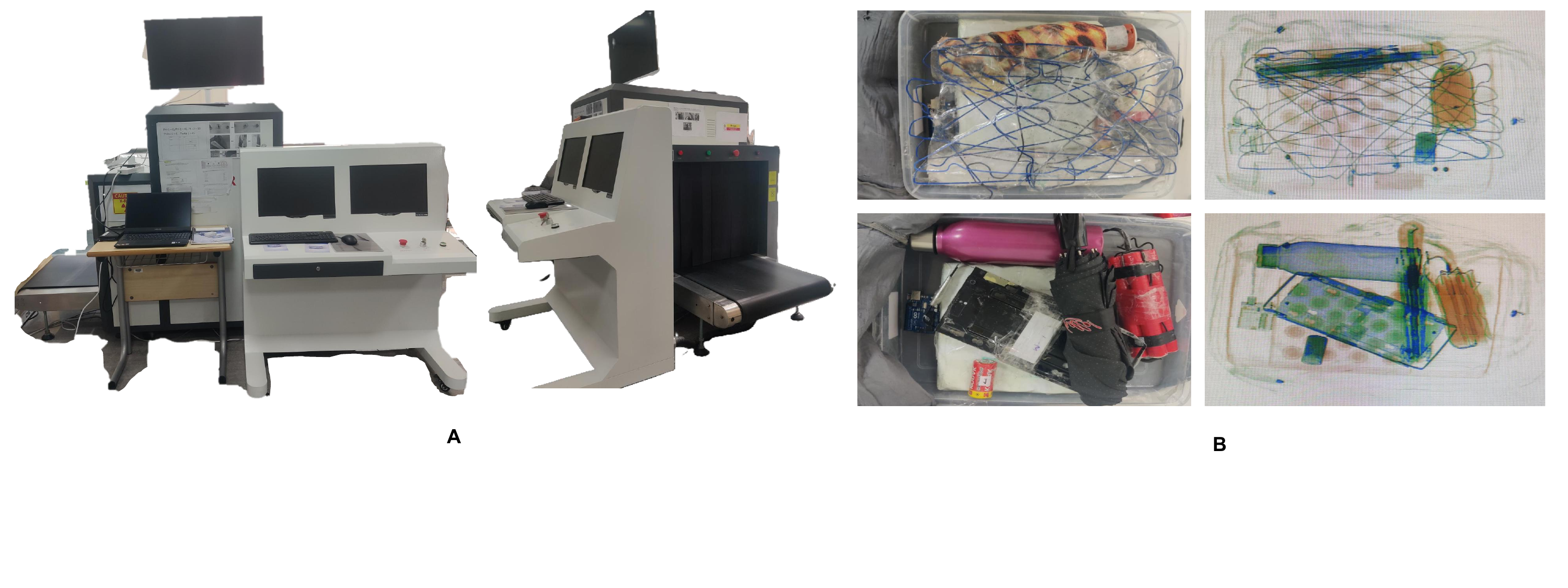}
\vspace{-1em}
\caption{\textbf{Acquisition setup and collection protocol.} 
(A) The dual-energy X-ray scanner used to acquire Falcon-X.
(B) Example baggage scenes and corresponding X-ray images showing diverse component layouts, superposition patterns, and concealment severity.
}
\label{fig:scanner}
\vspace{-1em}
\end{figure*}

\noindent All images are acquired using a dual-energy X-ray scanner (ANER K8065; 100--160\,kV, 0.4--1.2\,mA) in Fig. \ref{fig:scanner}(A). We collect baggage scans containing inert dismantled IED component samples drawn from a predefined functional taxonomy $\mathcal{C}$= \{\text{battery}, \text{detonator}, \text{main charge}\}.
To safely emulate explosive materials while preserving realistic attenuation characteristics, we use inert simulant powders selected to match density and effective atomic number under X-ray imaging \cite{eshetu2016inert}. 
We design a systematic collection protocol that reflects realistic concealment strategies by varying 
(i) spatial distribution of dismantled components within luggage, and
(ii) occlusion severity, ranging from fully visible arrangements to heavily concealed configurations (Fig. \ref{fig:scanner}(B)). This controlled design captures realistic concealment strategies while isolating functional composition as the primary risk factor. Following this protocol, we collect 7,000 base X-ray scans covering diverse component combinations, geometries, and clutter conditions (see Figs. \ref{fig:data_pipeline} and \ref{fig:tasksuite}).

\subsection{Samples of Counterfactual Synthetic Images}
\label{supplementarysubsec:synthetic}

\begin{figure*}[!t]
\centering
\includegraphics[width=0.99\linewidth,height=8cm]{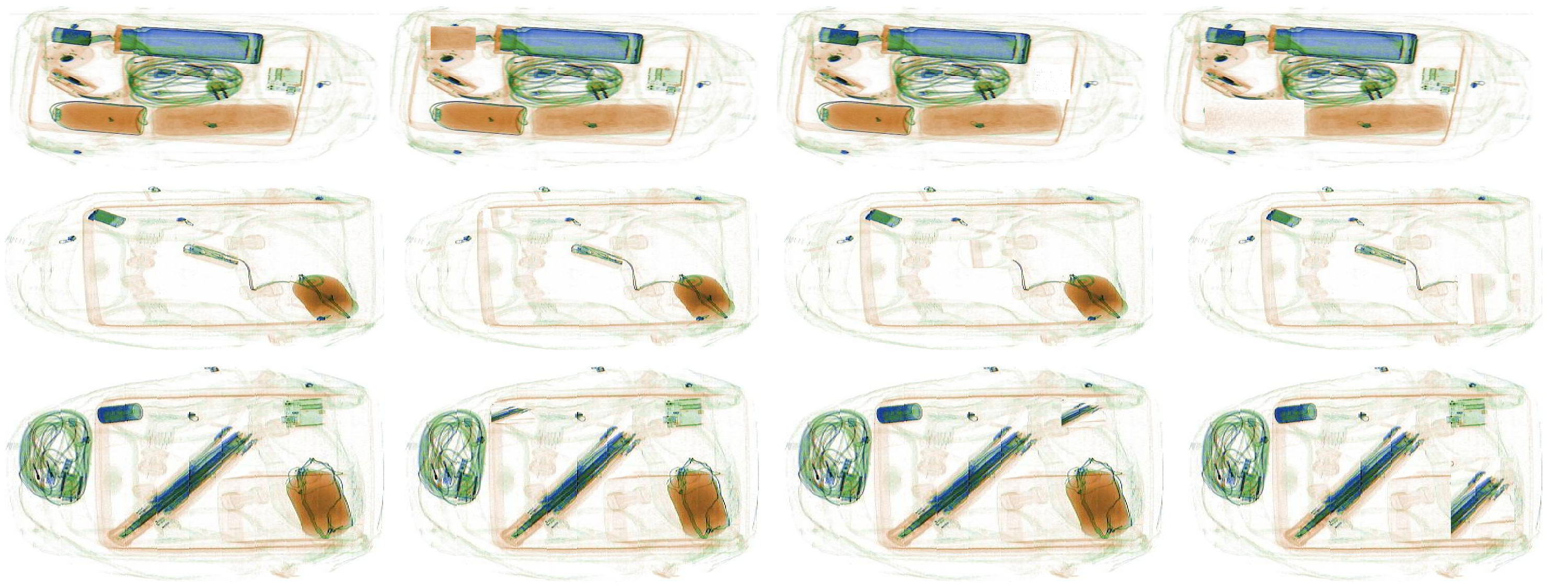}
\vspace{-1em}

\caption{
Examples of counterfactual synthetic samples used in Falcon-X. Starting from a real X-ray image (left), controlled variants are generated by removing one functional component using mask-guided inpainting. From left to right, each row shows: the original image, missing battery, missing detonator, and missing main charge variants. The background is filled using a texture sampled from surrounding regions of the same scan, preserving clutter statistics and visual realism while altering the compositional state of the scene.
}
\label{fig:synthetic_samples}
\vspace{-1em}
\end{figure*}

\noindent Fig. \ref{fig:synthetic_samples} presents representative examples of the counterfactual synthetic images used in Falcon-X. 
Starting from a real X-ray scan with instance-level masks, we generate semantically controlled variants by selectively removing one or more functional components from $\mathcal{C}=\{\text{battery},\text{detonator},\text{main charge}\}$ using mask-guided inpainting. 
The removed region is filled with background content sampled from the same image, which preserves local texture, clutter statistics, and overall X-ray appearance more faithfully than naive masking or zero-filling.

This process produces visually coherent counterfactual scenes in which the compositional state is altered while the surrounding baggage context remains largely unchanged. 
As a result, the model must reason over which components are present, which are missing, and how the modified configuration affects functional completeness and scene-level risk, rather than exploiting obvious synthetic artifacts. 
The generated samples cover complete assemblies, partial assemblies, and single-component configurations, enabling controlled supervision for missing-component identification, functional completeness estimation, and referring functional grounding.

\subsection{Quantitative Counterfactual Realism Check}
\label{supp:counterfactual_realism}
\vspace{-1em}
To verify that mask-guided counterfactuals do not introduce obvious visual artifacts, we compare them against held-out real X-ray images and a zero-fill removal baseline. 
We report feature-level distribution metrics (DINO-FID and DINO-KID), intensity-distribution shift (histogram EMD), and local boundary continuity around the edited mask region. 
As shown in Table \ref{tab:cf_realism}, mask-guided counterfactuals are substantially closer to the real-vs-real split than zero-fill removal across all metrics. 
This supports their use as controlled compositional variants while preserving realistic X-ray appearance and local mask-boundary consistency.
\vspace{-1.5em}
\begin{table}[!h]
\centering
\vspace{-1em}
\caption{Quantitative realism check for counterfactual images.}
\vspace{-0.1em}
\setlength{\tabcolsep}{10pt}
\renewcommand{\arraystretch}{0.8}
\scalebox{0.75}{
\begin{tabular}{l|cccc}
\hline
\rowcolor{Gray}
{\small \textbf{Image set}} 
& {\small \textbf{DINO-FID$\downarrow$}} 
& {\small \textbf{DINO-KID$\downarrow$}}
& {\small \textbf{Hist. EMD$\downarrow$}} 
& {\small \textbf{Boundary gap$\downarrow$}} 
\\
\hline

\small Real-vs-real split 
& 4.08 & 1.21 & 0.017 & 0.031 \\
\small Zero-fill removal 
& 32.74 & 19.86 & 0.094 & 0.486 \\
\rowcolor{gaincolor}
\small \textbf{Mask-guided counterfactuals} 
& 6.72 & 2.94 & 0.023 & 0.061 \\
\hline
\end{tabular}
}
\label{tab:cf_realism}
\vspace{-4em}
\end{table}

\section{Additional Details on Falcon-X Task Suite Generation}
\label{supplementary: Additional Details on FALCON-X task suite generation}
\vspace{-1em}

\subsection{Semi-Automated annotation}
\label{supplementary: Additional Details on FALCON-X task suite automated generation}


\noindent This section provides the detailed prompt template used to generate the Falcon-X task suite described in Sec. \ref{sec:tasks}. 
Task generation is performed from structured scene annotations rather than free-form image descriptions. 
Specifically, the prompt consumes image-level metadata, component-level attributes, and precomputed spatial relations derived from the annotated X-ray scan, and instructs the generator to produce a logically consistent set of multimodal tasks in a unified JSON format. 
This makes all generated captions, referring expressions, VQA pairs, functional presence labels, relational links, risk scores, and counterfactual variants remain grounded in the annotated scene and aligned with the compositional threat reasoning framework introduced in the main paper.

The template is intentionally constrained to prevent hallucination and enforce consistency between linguistic outputs and structured safety annotations. 
In particular, the generator is restricted to visible components only, must formulate uniquely resolvable grounding instructions, and must assign functional links and risk levels according to predefined compositional rules. 
The resulting JSON output is then used as the intermediate representation for constructing the final Falcon-X task suite, followed by verification and manual curation where necessary.

\begin{tcolorbox}
SYSTEM:
You are an expert security analyst specializing in X-ray threat analysis and structured functional reasoning.

You must generate tasks strictly grounded in the provided visual and spatial metadata.
Do NOT hallucinate unseen components.
Use only the listed components and their relationships.
All outputs must be logically consistent with the structured annotations.

------------------------------------------------------------
\end{tcolorbox}

\begin{tcolorbox}

USER INPUT:

[Image-Level Information]
- Image ID: \{$image-{id}$\}
- Number of detected components: \{N\}
- Global visual descriptors: \{tags\}

------------------------------------------------------------

[Component-Level Information]

For each detected component i:

Component {i}:
- Category: {battery | detonator | main charge}
- Bounding box: (x1, y1, x2, y2)
- Normalized center: (cx, cy)
- Area (pixels): {area}
- Mask shape: \{elongated | circular | irregular | compact\}
- Overlapping components: \{list\}
- Crop image: <attached>
- Binary mask: <attached>

------------------------------------------------------------

[Precomputed Spatial Relations]

For each pair (i, j):
- Center distance (normalized): $d-{ij}$
- IoU: iou-ij
- Relative position: \{left of | right of | above | below | overlapping\}
- Proximity: \{touching | near | far\}

------------------------------------------------------------

TASK:

Generate a structured Falcon-X task suite in STRICT JSON format
with the following fields:

\{
  "scene-caption": "...",

  "referring-segmentation": [
    \{"instruction": "...", 
    "target-component-id": i\}
  ],

  "vqa": [
    \{
      "question": "...",
      "answer": "...",
      "supporting-component-ids": [...]
    \}
  ],

  "functional-presence": \{
    "battery": 0/1,
    "detonator": 0/1,
    "main charge": 0/1
  \},

  "functional-links": [
    \{
      "pair": [i, j],
      "functionally-linked": 0/1,
      "justification": "..."
    \}
  ],

  "scene-risk-score": 0.0-1.0,
  "risk-explanation": "...",

  "counterfactual-variants": [
    \{
      "removed-component-id": i,
      "new-risk-score": 0.0-1.0,
      "explanation": "..."
    \}
  ]
\}

CONSTRAINTS:

1. Scene captions must describe only visible components.
2. Referring instructions must be uniquely resolvable from spatial metadata.
3. VQA questions must be answerable from structured annotations.
4. Functional links must be inferred using spatial proximity and component types.
5. Risk score guidelines:
   - 0.0–0.3: benign
   - 0.3–0.6: incomplete assembly
   - 0.6–1.0: high functional risk
6. Counterfactual variants must simulate removal of existing components only.
7. Do NOT introduce new objects.

Return JSON only. No explanations outside JSON.
\end{tcolorbox}

\subsection{Risk Label Generation and Verification}
\label{supp:risk_labels}

Falcon-X assigns a scene-level risk score $r\in[0,1]$ from structured annotations using component presence, type-level compatibility, and visual uncertainty caused by occlusion or concealment. 
Functional completeness measures whether required parts are present and compatible, whereas risk additionally reflects ambiguity in the visual evidence. 
Initial scores are generated using the rubric in Table \ref{tab:risk_rubric} and then verified by expert reviewers calibrated on representative benign, incomplete, ambiguous, and high-risk cases. 
Scores inconsistent with expert judgment are manually corrected, and ambiguous cases are resolved by expert adjudication. Expert verification accepted roughly about 72.3\% of generated labels without change and corrected 27.7\% labels.

\vspace{-1.5em}

\begin{table}[h]
\centering
\caption{Risk rubric used for Falcon-X label generation and expert verification.}
\vspace{-1em}
\small
\setlength{\tabcolsep}{4pt}
\begin{tabular}{p{0.18\linewidth}p{0.72\linewidth}}
\hline
\textbf{Risk range} & \textbf{Criteria} \\
\hline
$[0.0,0.3)$ 
& Low risk: required components are absent, isolated, or visually unsupported by compatible evidence. \\

$[0.3,0.6)$ 
& Medium risk: partial or ambiguous assembly; some components are present, but functional sufficiency is uncertain due to missing evidence, clutter, or occlusion. \\

$[0.6,1.0]$ 
& High risk: required components are present or strongly supported by visual evidence, and the configuration admits a plausible functional assembly. \\
\hline
\end{tabular}
\label{tab:risk_rubric}
\end{table}

\vspace{-2.5em}
\section{Qualitative Results}
\label{supplementary: Additional Qualitative results}

\begin{figure}[!h]
\vspace{-1em}
\centering
\includegraphics[width=0.8\linewidth, height=5cm]{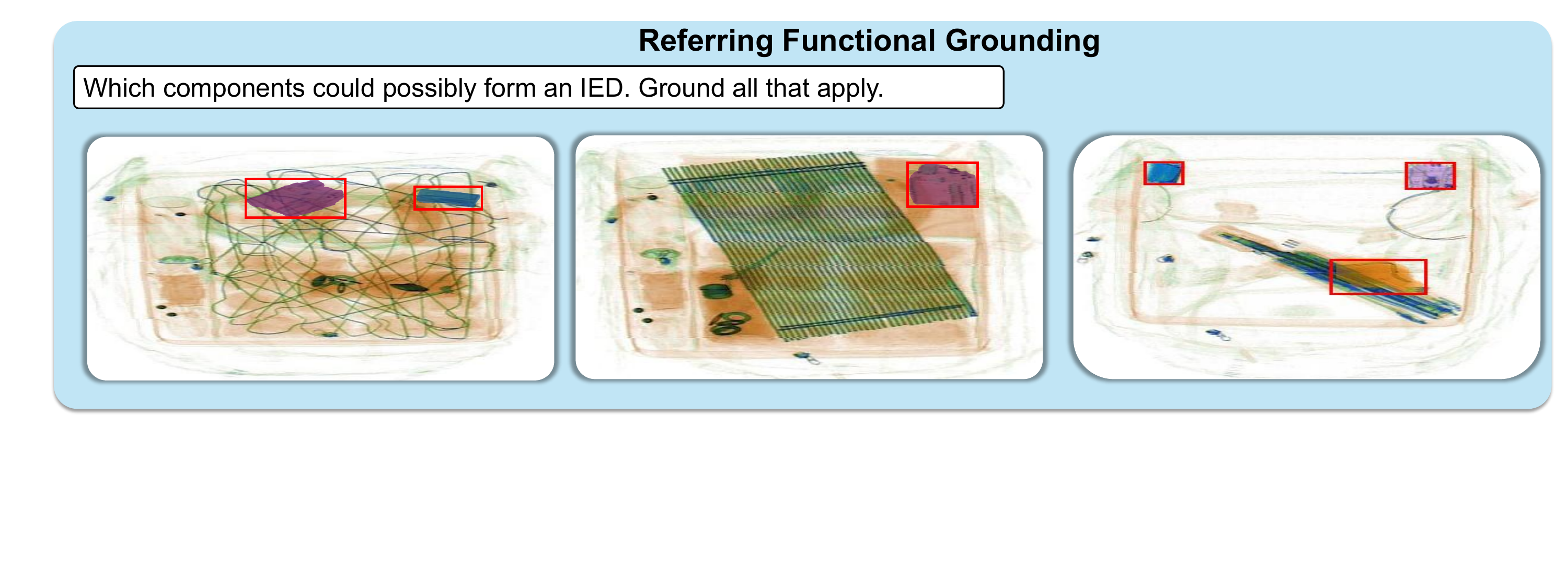}
\vspace{-4em}
\caption{\textbf{Qualitative examples of Falcon on RFG.} 
Given a functionally constrained query, the model grounds all visible components that could jointly participate in a potential IED assembly. 
}
\label{fig:rfg_qualitative}
\vspace{-1em}
\end{figure}

\noindent \textbf{RFG.}
Fig. \ref{fig:rfg_qualitative} presents qualitative examples of RFG on Falcon-X. 
Given the query, \emph{“Which components could possibly form an IED. Ground all that apply.”}, Falcon identifies and localizes all visible components that are functionally relevant to a potential assembly, rather than grounding a single object category in isolation. The examples show that the model can jointly select spatially separated components, including partially occluded instances under heavy X-ray clutter and superposition. 

\begin{figure*}[!h]
\vspace{-2em}
\includegraphics[width=0.5\linewidth,height=8cm]{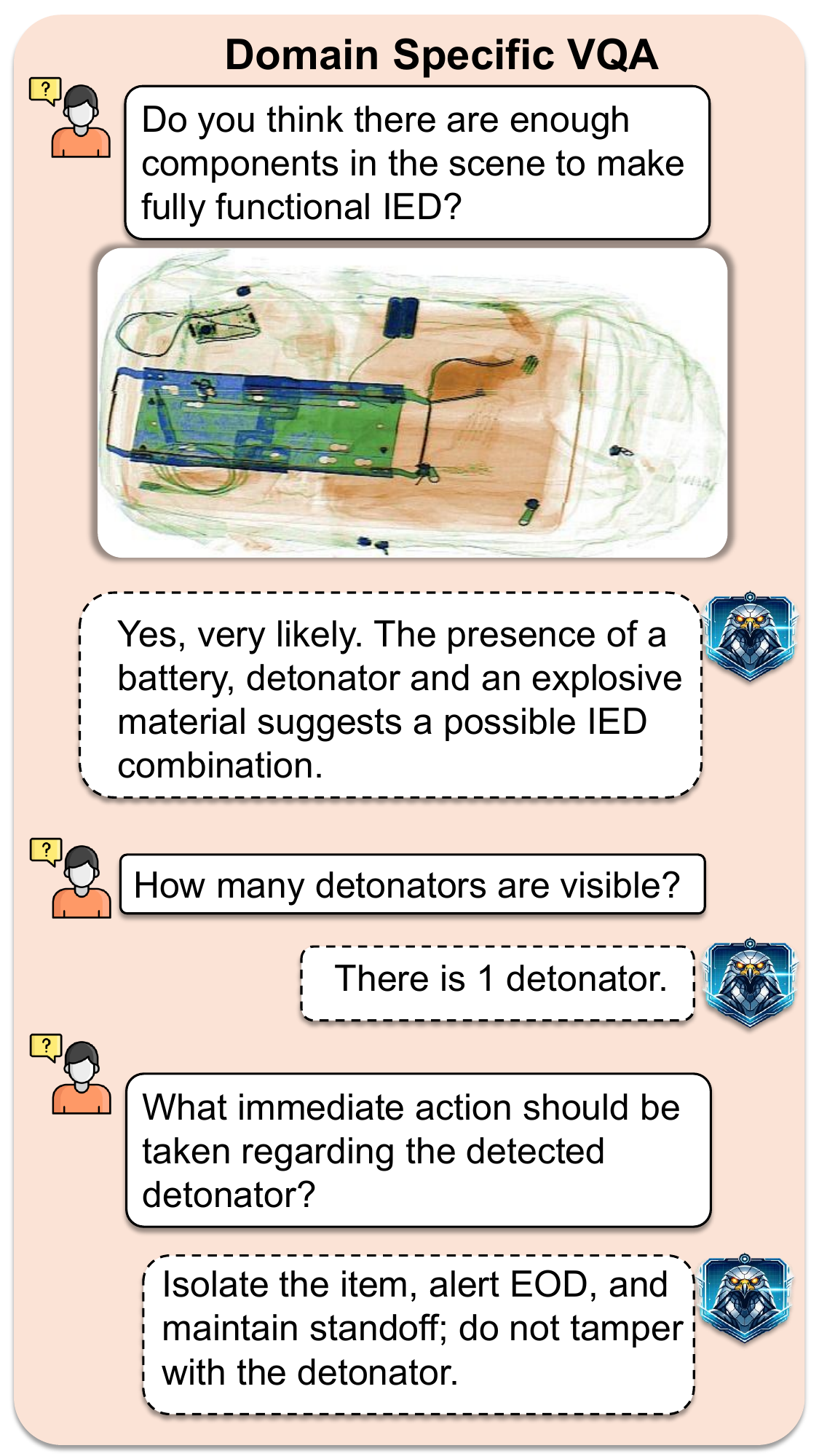}
\includegraphics[width=0.5\linewidth,height=8cm]{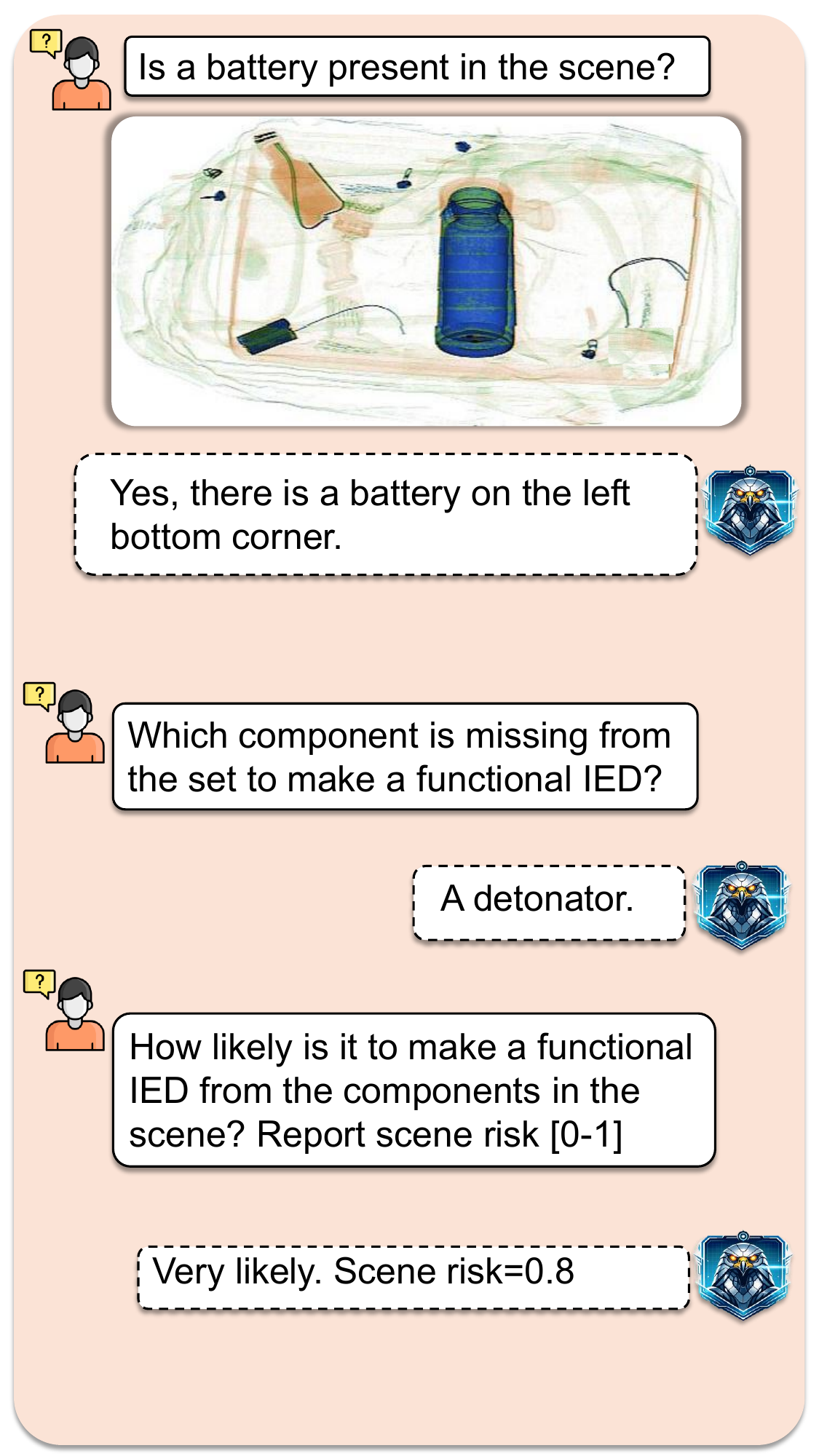}
\vspace{-2em}
\caption{
\textbf{Qualitative examples of Falcon on domain-specific VQA in X-ray imagery.} The examples illustrate multi-level question answering over compositional threat scenarios, including component presence, missing-component reasoning, scene-risk estimation, functional completeness, counting, and immediate response queries.
}
\label{fig:vqa}
\vspace{-4em}
\end{figure*}

\noindent \textbf{VQA.}
Fig. \ref{fig:vqa} shows representative qualitative examples of Falcon on domain-specific VQA. 
Beyond standard presence and counting questions, Falcon answers higher-level safety queries that require reasoning over the functional composition of the scene. 
In the left example, the model identifies the presence of a battery, infers that a detonator is missing, and assigns a high scene-risk score based on the visible configuration. 
In the right example, Falcon recognizes that the visible set is sufficient to plausibly form a functional IED, correctly counts the detonator, and generates an operationally appropriate response for immediate handling. Falcon’s structured safety conditioning yields visually grounded, relationally coherent answers for compositional threat reasoning.

\section{Cross-dataset Generalization} 
\label{supplementary: Cross-dataset evaluation}
\vspace{-1em}
\noindent To assess transfer beyond Falcon-X, we evaluate referring localization on STCray \cite{stingbee}, PIDray \cite{pidray}, and SIXray \cite{sixray} without further instruction tuning. 
For each dataset, referring-localization queries are generated from available annotations using the same protocol as Falcon-X detailed in Sec. \ref{subsec:annotation}. 

\noindent Table \ref{tab:crossdataset} shows that Falcon consistently outperforms Sting-Bee across all three datasets, with average gains of $+6.71$ cIoU and $+10.64$ mIoU. This suggests that Falcon's proposal and mask-aware grounding capture transferable X-ray structural cues beyond Falcon-X. 
\begin{table}[!h]
\centering
\vspace{-2em}
\caption{Cross-dataset evaluation results on referring localization.}
\vspace{-1em}
\setlength{\tabcolsep}{2pt}
\scalebox{0.69}{
\begin{tabular}{l|cc|cc|cc}
\hline
\rowcolor{Gray} {\small \textbf{Method}} & \multicolumn{2}{c}{\small \textbf{STCray}} & \multicolumn{2}{c}{\small \textbf{PIDray}} & \multicolumn{2}{c}{\small \textbf{SIXray}} \\
\cline{2-7}
\rowcolor{Gray}
& \small \textbf{cIoU} & \small \textbf{mIoU} & \small \textbf{cIoU} & \small \textbf{mIoU} & \small \textbf{cIoU} & \small \textbf{mIoU} \\

\hline
Sting-Bee
& 10.11 & 21.34
& 12.72 & 24.88
& 8.94 & 18.76 \\

\rowcolor{gaincolor} \textbf{Falcon}
& 16.60 $\Delta$\footnotesize{(\textcolor{gray}{\textbf{+6.49$\uparrow$}})}
& 32.12 $\Delta$\footnotesize{(\textcolor{gray}{\textbf{+10.78$\uparrow$}})}
& 20.45 $\Delta$\footnotesize{(\textcolor{gray}{\textbf{+7.73$\uparrow$}})}
& 36.31 $\Delta$\footnotesize{(\textcolor{gray}{\textbf{+11.43$\uparrow$}})}
& 14.85 $\Delta$\footnotesize{(\textcolor{gray}{\textbf{+5.91$\uparrow$}})}
& 28.47 $\Delta$\footnotesize{(\textcolor{gray}{\textbf{+9.71$\uparrow$}})} \\
\hline
\end{tabular}
}
\label{tab:crossdataset}
\vspace{-1em}
\end{table}
%







\section{Additional Details on Experimental Setup}
\label{supplementary: Additional Details on Experiment Setup}

\noindent We train Falcon in three stages: (1) detector fine-tuning, (2) structured vision-language alignment pretraining, and (3) instruction fine-tuning. 
Unless otherwise specified, all experiments are conducted with mixed-precision training on two NVIDIA A100 GPUs. 
Table \ref{tab:training_setup} summarizes the main optimization settings used in each stage.

\noindent \textbf{Stage 1: Detector fine-tuning.}
In the first stage, we fine-tune RF-DETR in a class-agnostic manner using the Falcon-X instance annotations, where all functional components are treated as foreground objects for proposal learning. 
Training is performed for 12 epochs with AdamW and a cosine learning-rate schedule. 
All images are resized to $448 \times 448$. 
This stage establishes the segmentation-aware perception module used to generate box and mask proposals for the later multimodal stages.

\noindent \textbf{Stage 2: Structured vision-language alignment.}
In the second stage, the detector is frozen and used only to provide region proposals and masks. 
The visual backbone and Vicuna-7B language model are also kept frozen, while the mask-aware projection layers and Structured Safety Adapter are optimized to align region-level X-ray features with language and structured safety targets. 
Training is performed for one epoch on the generated multimodal instruction set, using AdamW with a learning rate of $1\times10^{-4}$ and warmup ratio 0.03. 
At this stage, the model learns to map segmentation-aware proposals into component presence, functional links, scene-level risk, and grounded textual responses.

\noindent \textbf{Stage 3: Instruction fine-tuning.}
Starting from the Stage-2 checkpoint, we enable LoRA adapters in the LLM attention layers while keeping the detector frozen. 
The projection layers and Structured Safety Adapter remain trainable. 
This stage uses a smaller learning rate of $1\times10^{-5}$ and is run for one epoch on the same instruction corpus. 
The purpose of this stage is to improve instruction following, response quality, and consistency between generated language and the inferred structured safety state.

\noindent \textbf{Implementation details.}
We use DINOv2-L/14 as the image encoder and Vicuna-7B as the language backbone. 
For each image, RF-DETR produces up to 300 proposals before score filtering and non-maximum suppression, after which the top 100 regions are retained. 
ROI-aligned and mask-pooled features are fused into region embeddings and passed to the Structured Safety Adapter, which predicts component presence, pairwise functional links, and scene risk. 
These predictions are then converted into seven structured safety tokens and concatenated with image, region, and text tokens before decoding.

\noindent \textbf{Training data.}
Stage 1 is trained on Falcon-X instance-level annotations, comprising 16,580 annotated component instances. 
Stages 2 and 3 use the generated Falcon-X multimodal instruction data, which contains 442,287 training instructions spanning captioning, VQA, referring grounding, functional completeness, missing-component identification, and risk reasoning tasks. 
All counterfactual variants derived from the same base image are kept within the same split to avoid train-test leakage.

\noindent \textbf{Efficiency.}
Under this setup, detector fine-tuning requires approximately 6 hours, while Stage 2 and Stage 3 require approximately 22 and 18 hours, respectively. 
This staged design keeps the most expensive vision module fixed during multimodal adaptation and enables efficient domain-specific training with limited compute.

\begin{table*}[!h]
\vspace{-1em}
\centering
\caption{Training configuration across the three stages.}
\vspace{-1em}
\label{tab:training_setup}
\small
\resizebox{0.99\textwidth}{!}{%
\begin{tabular}{lccc}
\hline
\textbf{Configuration} 
& \textbf{Stage 1} 
& \textbf{Stage 2} 
& \textbf{Stage 3} \\
\hline
Optimizer 
& AdamW
& AdamW 
& AdamW \\

Epochs 
& 12 
& 1 
& 1 \\

Per-device batch 
& 64 
& 4 
& 4 \\

Learning rate 
& $1\times10^{-4}$ 
& $1\times10^{-4}$ 
& $1\times10^{-5}$ \\

Weight decay 
& $1\times10^{-4}$ 
& 0 
& 0 \\

LR schedule 
& cosine
& cosine 
& cosine \\

Warmup ratio 
& None 
& 0.03 
& 0.03 \\

Precision 
& mixed precision 
& fp16 (+ tf32) 
& bf16 (+ tf32) \\

Resolution 
& $448\times448$
& $448\times448$ 
& $448\times448$ \\

Training time (hrs) 
& 6
& 22 
& 18 \\

Training samples 
& 16,580 instances
& 442,287 instructions 
& 442,287 instructions \\

LLM update mode 
& frozen
& frozen 
& LoRA (r=16, $\alpha$=32, dropout=0.05) \\

\hline
\end{tabular}
}
\vspace{-2em}
\end{table*}
\section{Additional Ablation study}
\label{supplementary:Additional Ablation study}
\subsection{Ablations Beyond RFG}

To complement the RFG prediction-head ablation in the main paper, we evaluate Falcon under progressively stronger adaptation strategies across additional Falcon-X tasks (Table \ref{tab:falcon_finetuning_ablation}). 
This ablation separates the effect of generic multimodal adaptation from explicit functional-state supervision. 
All variants use the same task suite, while progressively enabling projector tuning, instruction tuning, LoRA adaptation, and SSA multi-task supervision.

\begin{table}[!h]
\centering
\vspace{-1em}
\caption{Falcon adaptation strategy ablation under different fine-tuning strategies.}
\setlength{\tabcolsep}{8pt}
\scalebox{0.75}{
\begin{tabular}{l|cc|cc|ccc}
\hline
\rowcolor{Gray}
{\small \textbf{Strategy}} & \multicolumn{2}{c|}{\small \textbf{Semantic Grounding}}& \multicolumn{2}{c|}{\small \textbf{RFG}} & \multicolumn{3}{c}{\small \textbf{SR}}\\
\cline{2-3}
\cline{6-8}
\rowcolor{Gray} & \small \textbf{CPC} &\small \textbf{MCI} &   &  &  \small \textbf{FC} & \small \textbf{SRL} & \small \textbf{CLR}\\
\cline{2-8}

\rowcolor{Gray} & {\small Acc$\uparrow$} & {\small Acc$\uparrow$}
& {\small cIoU$\uparrow$}
& {\small mIoU$\uparrow$}
& 
& {\small MAE$\downarrow$}
&  \\

\hline
Prompt-only / zero-shot
& 54.18 & 21.74 & 12.63 & 19.85 & 0.164 & 0.286 & 0.271 \\

Projector-only tuning
& 90.72 & 83.45 & 33.94 & 44.81 & 0.061 & 0.094 & 0.073 \\

Instruction tuning
& 94.31 & 88.62 & 38.27 & 50.36 & 0.052 & 0.078 & 0.056 \\

LoRA adaptation
& 96.84 & 94.18 & 44.63 & 60.71 & 0.031 & 0.047 & 0.022 \\

\rowcolor{gaincolor}
Falcon: LoRA + SSA multi-task
& \textbf{98.1} & \textbf{94.75}
& \textbf{50.45} & \textbf{69.58}
& \textbf{0.017} & \textbf{0.020} & \textbf{0.005} \\

\hline
\end{tabular}
}
\label{tab:falcon_finetuning_ablation}
\vspace{-2em}
\end{table}

\subsection{Backbone Control Ablation}
\label{supp:backbone_control}

To isolate the contribution of the Structured Safety Adapter (SSA) from the choice of LLM backbone, we replace Vicuna-7B with a Qwen3.5-VL backbone while keeping the perception module, region/mask interfaces, SSA module, training data, and evaluation protocol fixed. 
As shown in Table \ref{tab:backbone_control}, changing the backbone without SSA yields only modest improvements. 
In contrast, enabling SSA produces substantially larger gains for both backbones across RFG and functional/safety reasoning metrics. 
This indicates that Falcon's improvement primarily comes from the explicit structured safety state rather than from the specific Vicuna-7B backbone.
\vspace{-1em}
\begin{table}[!h]
\centering
\caption{Backbone-control ablation. 
We compare Vicuna-7B and Qwen3.5-VL with and without SSA while keeping the remaining Falcon components fixed. 
RFG is reported with cIoU/mIoU; FC, SRL, and CLR are reported as MAE.}
\setlength{\tabcolsep}{4pt}
\scriptsize
\begin{tabular}{l|c|cc|ccc}
\hline
\rowcolor{Gray}
\textbf{Backbone} 
& \textbf{SSA} 
& \multicolumn{2}{c|}{\textbf{RFG}} 
& \multicolumn{3}{c}{\textbf{Functional/Safety}}\\
\cline{3-7}
\rowcolor{Gray} 
&  
& \textbf{cIoU}$\uparrow$ 
& \textbf{mIoU}$\uparrow$ 
& \textbf{FC}$\downarrow$ 
& \textbf{SRL}$\downarrow$ 
& \textbf{CLR}$\downarrow$\\
\hline

Vicuna-7B 
& \xmark 
& 35.82 & 43.43 & 0.046 & 0.081 & 0.054 \\

Qwen3.5-VL 
& \xmark 
& 36.91 & 45.72 & 0.044 & 0.078 & 0.052 \\

\hline

Vicuna-7B 
& \cmark 
& 50.45 & 69.58 & 0.017 & 0.020 & 0.005 \\

Qwen3.5-VL 
& \cmark 
& \textbf{51.07} & \textbf{70.14} & \textbf{0.016} & \textbf{0.019} & \textbf{0.004} \\

\hline
\end{tabular}
\label{tab:backbone_control}
\end{table}

\vspace{-1em}

\subsection{Failure Analysis of Fine-tuned VLMs}
\label{supp:vlm_failure_analysis}

To better characterize model errors on Falcon-X, we categorize failures into component-level and compositional reasoning errors. 
\textit{Component miss} denotes failure to detect or mention a visible required component; 
\textit{grounding mismatch} denotes incorrect spatial grounding; 
\textit{missing-part error} denotes incorrect identification of absent components; 
\textit{link error} denotes incorrect functional compatibility prediction; 
\textit{risk inconsistency} denotes safety assessments inconsistent with the predicted functional state; 
and \textit{hallucination} denotes references to unsupported components or relations.

Table \ref{tab:vlm_failure_analysis} shows that fine-tuned VLMs often adapt to X-ray appearance and achieve low component-miss rates, but still exhibit substantially higher link-error and risk-inconsistency rates. 
This indicates that Falcon-X exposes a compositional reasoning bottleneck beyond component recognition or segmentation. 
Falcon reduces these failures by conditioning generation on an explicit structured safety state.
\vspace{-1.5em}
\begin{table}[!h]
\centering
\caption{Failure analysis of fine-tuned VLMs on Falcon-X. 
Entries denote failure rates (\%) for component recognition, spatial grounding, missing-part reasoning, functional-link prediction, risk consistency, and hallucination.}
\setlength{\tabcolsep}{3.5pt}
\renewcommand{\arraystretch}{0.85}
\scalebox{0.62}{
\begin{tabular}{l|cccccc}
\hline
\rowcolor{Gray}
\textbf{Method} 
& \textbf{Comp. miss} 
& \textbf{Grounding mismatch}
& \textbf{Missing-part error} 
& \textbf{Link error} 
& \textbf{Risk inconsistency} 
& \textbf{Hallucination} \\
\hline
LISA \cite{lisa} 
& 8.7 & 23.4 & 32.8 & 41.5 & 38.2 & 10.6 \\
Sting-Bee \cite{stingbee}  
& 3.1 & 18.9 & 13.7 & 36.8 & 31.5 & 6.4 \\
Groma \cite{groma} 
& 2.8 & 15.1 & 8.9 & 29.6 & 25.4 & 4.8 \\
\rowcolor{gaincolor}
\textbf{Falcon} 
& \textbf{2.4} & \textbf{7.6} & \textbf{6.2} & \textbf{7.8} & \textbf{5.9} & \textbf{3.1} \\
\hline
\end{tabular}
}
\label{tab:vlm_failure_analysis}
\end{table}

\vspace{-2em}

\subsection{Standard vs. Functional Grounding Trade-off}
\label{supp:standard_functional_tradeoff}

Falcon prioritizes function-aware grounding over exhaustive panoptic coverage. To diagnose its weaker PS/RPS performance, we tested proposal recall, visual encoder freezing, and the structured safety bottleneck. 
Increasing proposal recall from $92.6\%$ to $97.8\%$ and partially unfreezing the visual encoder changed PS/RPS by less than $1$ cIoU, suggesting that proposal quality and freezing are not the main causes. 
In contrast, relaxing the structured bottleneck recovered part of the standard grounding drop ($+5.27$ PS cIoU and $+3.99$ RPS cIoU), but reduced RFG by $7.39$ cIoU and $10.78$ mIoU. 
This indicates that the trade-off is mainly induced by Falcon's structured functional interface, which favors safety-relevant component grounding over generic dense coverage.

\section{Future Work}
\label{supplementary: Future Work}
\noindent While Falcon-X establishes a benchmark for compositional threat reasoning in single-view X-ray imagery, several important directions remain open. First, an immediate extension is \emph{multi-view and multi-context reasoning}. Real screening is often limited by occlusion and superposition in a single scan, and recent dual-view X-ray studies show that complementary views improve recognition under heavy overlap \cite{dvxray}. A natural next step is to extend Falcon from single-image inference to reasoning over multiple synchronized views or related scans, where evidence for a functional assembly may be distributed across images rather than visible in one scene. This would extend compositional threat reasoning from intra-image relation modeling to cross-image evidence aggregation.

\noindent Second, Falcon currently operates on a fixed functional taxonomy of dismantled IED components. Future work should generalize this formulation to \emph{open-vocabulary and larger modular threats}, enabling reasoning over unseen components, substitutes, and partially novel assemblies. Recent progress in open-vocabulary X-ray detection \cite{raxo} suggests that moving beyond closed-set categories is both feasible and important for real-world deployment. Extending Falcon in this direction would enable evaluation under more realistic open-world conditions.

\noindent The third promising direction is to strengthen the structured reasoning layer itself. Although Falcon introduces an explicit safety state, future models could incorporate \emph{neuro-symbolic constraints}, graph-based inference, or self-correcting logical modules to better enforce consistency between component presence, pairwise compatibility, and scene-level risk. This is particularly relevant because recent work in visual grounding \cite{naiver,neursymb} shows that compositional and relational reasoning remain major weaknesses of current vision-language models.

\noindent Finally, future deployment-oriented work should emphasize \emph{uncertainty-aware decision support}. In safety-critical screening, accurate prediction alone is insufficient; systems must also recognize when evidence is incomplete, ambiguous, or unreliable \cite{vilu}. This motivates extensions of Falcon toward calibrated risk estimation, selective abstention, and confidence-aware human-AI collaboration, where uncertain cases are escalated for human review rather than forced into overconfident predictions. Advancing from single-view structured reasoning to multi-context, open-world, and uncertainty-aware threat analysis is a key next step for multimodal AI in security screening.



\end{document}